\documentclass{article}
\usepackage[numbers]{natbib}

 \usepackage[preprint]{neurips_2026}

\usepackage[utf8]{inputenc} % allow utf-8 input
\usepackage[T1]{fontenc}    % use 8-bit T1 fonts
\usepackage{hyperref}       % hyperlinks
\usepackage{url}            % simple URL typesetting
\usepackage{booktabs}       % professional-quality tables
\usepackage{amsfonts}       % blackboard math symbols
\usepackage{nicefrac}       % compact symbols for 1/2, etc.
\usepackage{microtype}      % microtypography
\usepackage{xcolor}         % colors
\usepackage{graphicx}
\usepackage{subcaption}
\usepackage{multirow}
\usepackage{rotating}
\usepackage{amsmath}
\usepackage{amssymb}
\usepackage{wrapfig}

\title{What Cohort INRs Encode and Where to Freeze Them}

\author{%
Vasiliki Sideri-Lampretsa$^{1,2}$ \quad Sophie Starck$^{1,2}$ \quad Robbie Holland$^{5,6}$ \\ \quad \textbf{Julian McGinnis}$^{1,2,4}$ \quad
\textbf{Daniel Rueckert}$^{1,2,3,4}$ \\
$^1$Chair of AI in Healthcare and Medicine, Technical University of Munich,
Germany \\
$^2$TUM University Hospital, Munich, Germany \quad $^3$Imperial College London \\ 
$^4$Munich Center for Machine Learning \\
$^5$Stanford Center for Artificial Intelligence in Medicine and Imaging, \\Stanford
University, Stanford, CA, USA \\
$^6$Department of Radiology, Stanford University, Stanford, CA, USA
}

\begin{document}

\maketitle

% -----------------------------------------------------------------------
\begin{abstract}
Reusing the early layers of cohort-trained INRs as initialization for new signals has been shown to accelerate and improve signal fitting, yet it remains unclear which 
layers of the shared encoder learn transferable representations and what those representations encode. 
We address both questions for two standard backbones, SIREN and Fourier-feature MLPs (FFMLP). 
First, sweeping the freeze depth across the shared encoder at test time, we find that the optimum coincides with the layer of highest weight stable rank. 
Moreover, freezing at this depth matches or improves on the standard fine-tuning recipe across all our experiments.
Second, identifying which layer transfers does not characterize what that layer encodes.
To address this we adopt sparse autoencoders (SAEs), the dominant tool in mechanistic interpretability, and present the first SAE decomposition of INR activations into sparse dictionary atoms.
Interestingly, SIREN and FFMLP achieve comparable cohort-fitting quality, but learn qualitatively different dictionaries.
Cohort SIREN's atoms are localized, tiling the coordinate plane such that each atom fires in a confined region independent of cohort content. 
Cohort FFMLP's atoms are image-spanning, tracing the contours of memorized cohort signals. 
Single-atom ablations confirm causal use of these dictionaries: a single FFMLP atom out of 4096 can drop PSNR by up to 10.6 dB across the image, while SIREN ablations remain confined to where the atom fires.
Together, these results give the first mechanistic account of what transfers in cohort-trained INRs and turn their activations into inspectable dictionary atoms. 
These tools open a path towards characterizing what INRs encode and towards architectures designed for generalization rather than memorization.
We plan to release our code upon acceptance.
\end{abstract}

% -----------------------------------------------------------------------
\section{Introduction}
\label{sec:intro}

Implicit neural representations (INRs) are coordinate-based multi-layer perceptrons (MLPs) that learn a continuous, functional approximation of signals. 
Popularized by their usage in NeRFs~\cite{mildenhall2021nerf}, INRs have demonstrated their utility across diverse signal modalities including images~\cite{tancik2020fourier,sitzmann2020implicit,mehta2021modulated,vyas2025learning,vyas2026surprising}, scenes~\cite{mildenhall2021nerf,barron2021mip}, shapes~\cite{park2019deepsdf,gropp2020implicit,davies2020effectiveness}, videos~\cite{chen2021nerv,chen2022videoinr}, medical data~\cite{mcginnis2023single,sideri2024sinr,friedrich2025medfuncta,wolterink2022implicit}, and signal compression~\cite{dupont2021coin,maiya2023nirvana,strumpler2022implicit,zhang2021implicit}. 
A key distinction of INRs is that each network fits a single signal, with learned features finely tuned to that signal's morphology, and every new signal requires optimization from scratch. 
Reducing this per-signal cost, by reusing structure learned from related signals, would significantly broaden the applicability of INRs.

\begin{figure*}[t]
    \centering
    \includegraphics[width=\textwidth]{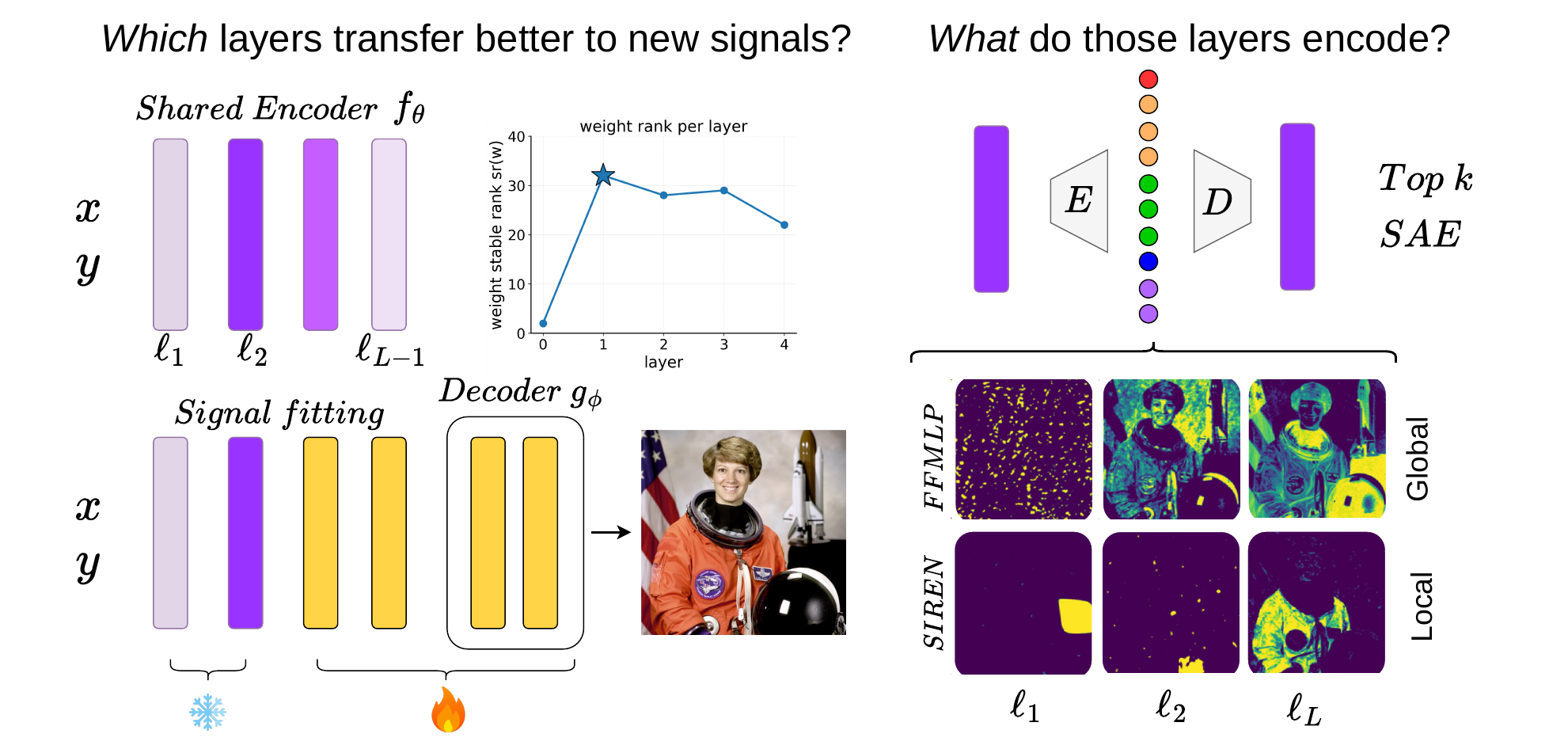}
    \caption{
        \textbf{Which layers transfer, and what do they encode?}
        \textbf{Left:} the optimal freeze depth $\tau^\star$ in cohort-trained INRs coincides with the layer of peak weight stable rank in the shared encoder ($\tau^\star = 1$ for cohort SIREN, shown).
        \textbf{Right:} sparse autoencoders recover qualitatively different dictionaries: SIREN atoms are localized, tiling the coordinate plane.
        FFMLP atoms are image-shaped and fire diffusely, tracing memorized cohort content.
        }
    \label{fig:teaser}
\end{figure*}

A recent line of work, STRAINER~\citep{vyas2025learning,vyas2026surprising}, addresses this by training a \emph{cohort} of INRs that share early encoder layers but maintain per-signal decoders. 
Once trained, the shared encoder serves as initialization for new signals.
STRAINER's test-time protocol updates every layer, so pretrained structure and test-time fitting cannot be separated, leaving per-layer transferability unmeasured.
This raises two questions: \emph{which} layers carry transferable representations, and \emph{what} do those representations encode?
Drawing inspiration from the pretrained vision encoder literature, where frozen-feature analysis has localized transferable representations to intermediate rather than output layers~\citep{bolya2025perception,el2024scalable,rajasegaran2025empirical}, we sweep the test-time freeze boundary across all encoder layers of a cohort-trained INR.
Whether this pattern holds for cohort-trained INRs has not been studied.
The two dominant backbones, SIREN~\citep{sitzmann2020implicit} and Fourier-feature MLPs (FFMLP)~\citep{tancik2020fourier}, may not localize transferable structure at the same depth since they encode coordinate position through fundamentally different mechanisms (sinusoidal activations at every layer versus a random Fourier projection followed by ReLU activations).
Figure~\ref{fig:teaser} summarizes our two findings.

To address the \emph{which} question, we train cohort INRs on three datasets (CelebA-HQ, OASIS-MRI, Kodak) with two backbones (SIREN and FFMLP), and evaluate whether frozen features transfer from each source to each of the three target distributions. 
Sweeping the freeze boundary at test time, we find that the optimal freeze depth coincides with the layer with the peak stable rank of the weights in the cohort-trained encoder~\citep{ramasinghe2022beyond,mcginnis2025optimizing}.
The weight stable rank is computable from the encoder alone, with no target dataset access or hyperparameter search.
% The peak occurs at the first trainable layer that maps the rank-2 coordinate input into a high-dimensional representation, the *rank lift*. 
Although the two backbones encode coordinate position differently, both transfer best when frozen up to the first high-rank layer. 
In FFMLP this layer is the first trainable layer after the fixed Fourier projection, and its features reflect the Fourier basis~\cite{padmanabhan2024explaining}. 
In SIREN, where no such fixed basis exists, the first high-rank layer itself can be viewed as an emergent positional embedding learned from the data.
To address the \emph{what} question, we adopt sparse autoencoders (SAEs)~\citep{bricken2023monosemantic,cunningham2023sparse,gao2024scaling}, the dominant tool for interpreting hidden activations in language and vision-language models. 
We apply SAEs to trained INRs, learning an overcomplete dictionary that decomposes activations into discrete atoms.
Despite training on the same cohort to comparable quality, SIREN and FFMLP recover qualitatively different dictionaries (Figure~\ref{fig:atom-gallery}). 
SIREN's atoms tend to be localized, tiling the coordinate plane independently of cohort content. 
FFMLP's atoms tend to be image-shaped, tracing memorized cohort signals. 
Single-atom ablations confirm the dichotomy is causal: SIREN ablations stay confined to where each atom fires, while FFMLP ablations propagate across the entire image, with single atoms causing PSNR drops of up to 10.6 dB (Fig.~\ref{fig:atom-ablation}).
We make the following contributions:
\begin{enumerate}
    \item We establish that frozen features transfer across cohort-trained INRs. 
    The optimal freeze depth coincides with the layer with the highest weight stable rank in the encoder. 
    Freezing at this depth matches or improves on the standard fine-tuning recipe across all our experiments.
    \item We present the first SAE decomposition of INR activations into sparse atoms, showing that SIREN and FFMLP recover qualitatively different dictionaries: localized atoms for SIREN vs. image-spanning atoms tracing memorized cohort content for FFMLP.
    \item We demonstrate, through single-atom ablations, that these dictionaries causally drive the network's output: SIREN ablations stay local to where the atom fires. FFMLP ablations propagate across the entire image, with single atoms causing PSNR drops of up to 10.8 dB.
\end{enumerate}
% -----------------------------------------------------------------------
\section{Related Work}
\label{sec:related}

\paragraph{Beyond the single-signal INR regime.}
Recent work tackles the cost of training INRs from scratch by learning priors over INR weights. Hypernetwork~\citep{sitzmann2020implicit,chen2022transformers} and meta-learning~\citep{tancik2021learned,sitzmann2020metasdf,lee2021meta} approaches map a code or initialization to target weights. 
Auto-decoder approaches~\citep{park2019deepsdf,dupont2022data,friedrich2025medfuncta,mehta2021modulated} learn a shared network conditioned on per-signal codes or modulations, and a separate line embeds a learned prior into the INR itself by training on related signals~\citep{shen2022nerp,stolt2023nisf,dannecker2024cina,lanzendorfer2023siamese}. 
Most relevant to our setup, STRAINER~\citep{vyas2025learning} trains a cohort with shared early layers and per-signal decoders, then uses the trained encoder as initialization for new signals, while its concurrent extension~\citep{vyas2026surprising} pretrains on uniform noise. IPC~\citep{kim2023generalizable} instead trains a transformer hypernetwork to predict per-instance weights for one MLP layer while sharing the rest. 
In this work, we adopt STRAINER's training unchanged and introduce a test-time freezing protocol that identifies which trained layers carry transferable representations, a question this prior work does not address.

\paragraph{Frozen-feature analysis and transfer learning.}
Transfer learning leverages source-distribution knowledge to improve a related target task~\citep{pan2009survey,caruana1997multitask,panigrahi2020survey}. 
Deep features are broadly reusable, with lower layers transferring more readily than task-specific upper ones~\citep{yosinski2014transferable,donahue2014decaf}, an insight underlying the pretrain-then-finetune paradigm dominant in vision~\citep{he2022masked,radford2021learning} and language~\citep{devlin2019bert,raffel2020exploring}. 
Within this paradigm, layer-wise frozen-feature analysis has consistently localized the most transferable features to intermediate layers, with output layers specializing toward the pretraining objective. The pattern recurs across autoregressive image and video modeling~\citep{el2024scalable}, contrastive vision-language pretraining~\citep{bolya2025perception}, self-supervised features~\citep{oquab2023dinov2}, and diffusion-based generation~\citep{yu2024representation}.
We adopt the same probe in the cohort-INR setting and find that the optimal freeze depth is identifiable a priori from the encoder's peak weight stable rank, with no target data or hyperparameter sweep required.

\paragraph{Characterizing what trained INRs encode.}
A parallel line of work characterizes the internal structure of trained INRs.
Analytical characterizations describe the input-output map directly. 
Spectral analyses~\citep{tancik2020fourier,basri2020frequency,rahaman2019spectral} explain the low-frequency bias of vanilla MLPs and motivate positional embeddings. 
Yüce et al.~\citep{yuce2022structured} express the coordinate-MLP function as a structured dictionary fixed at initialization, a view also developed in~\citep{benbarka2022seeing} from the Fourier-series perspective.
Empirical characterizations probe the trained network. 
Ramasinghe and Lucey~\citep{ramasinghe2022beyond} tie expressiveness to the singular-value distribution of hidden representations, and McGinnis et al.~\citep{mcginnis2025optimizing} recast spectral bias as stable-rank degradation during training.
The closest empirical attempts to inspect what trained INRs encode are SplineCAM~\citep{humayun2023splinecam}, which computes the input-space partition of ReLU networks, and XINC~\citep{padmanabhan2024explaining}, which attributes pixels to individual neurons. 
Both inherit the neuron basis as given and report what each neuron does in isolation, leaving unrecovered the directions the network actually uses to compose its outputs.
Our SAE-based decomposition is complementary to all of the above. 
It operates on intermediate activations rather than the input-output map, and it recovers an overcomplete basis the network engages rather than reading off the neuron basis. No prior work recovers a sparse decomposition of INR activations to expose the dictionary that a trained network actually uses.

\paragraph{Sparse autoencoders and dictionary learning.}
Sparse autoencoders (SAEs) decompose the activations of trained neural networks into interpretable, causally engaged features, building on classical sparse coding and dictionary learning~\citep{olshausen1997sparse,elad2006image}. 
Overcomplete sparse decompositions of transformer hidden states mitigate polysemanticity and recover causally meaningful features in language models~\citep{bricken2023monosemantic,cunningham2023sparse,templeton2024scaling} and vision models~\citep{gorton2024missing}, with TopK sparsity~\citep{makhzani2013k,gao2024scaling} preferred over $\ell_1$ regularization to avoid magnitude bias. 
SAE features form causal circuits when ablated or steered~\citep{marks2024sparse}.
Compression-oriented work like COIN~\citep{dupont2021coin}, Strümpler et al.~\citep{strumpler2022implicit}, and SiNR~\citep{jayasundara2025sinr} applies sparse decomposition to INR weights for storage, distinct from our goal of decomposing activations to interpret a trained INR.
To our knowledge, SAEs have not previously been applied to the hidden activations of implicit neural representations.

% -----------------------------------------------------------------------
\section{Methods}
\label{sec:methods}

We study cohort-trained INRs through two diagnostics on the trained encoder. To localize the transferable layer, we sweep a freeze boundary at test time and find the optimum coincides with the layer of peak weight stable rank, a quantity computed from the encoder alone (Sec.~\ref{sec:method-rank}). To characterize what that layer encodes, we fit a sparse autoencoder to its activations, recovering an overcomplete basis of atoms that supports causal intervention (Sec.~\ref{sec:method-sae}). Both diagnostics build on the cohort training and freezing protocol of Sec.~\ref{sec:method-setup}.

\subsection{Cohort INR transfer}
\label{sec:method-setup}

An INR parameterizes a signal as a continuous function $f : \mathcal{X} \to \mathbb{R}^c$ over a coordinate domain $\mathcal{X} \subseteq \mathbb{R}^m$. 
For images, $\mathcal{X} = [-1, 1]^2$ and $f : \mathcal{X} \to \mathbb{R}^3$ maps 2D coordinates to RGB. 
We implement $f$ as an MLP with $L$ hidden layers of width $d$ followed by a linear output projection:
\begin{equation}
\label{eq:inr-forward}
\mathbf{h}_0 = \sigma(W_0 \gamma(\mathbf{x}) + \mathbf{b}_0), \quad \mathbf{h}_\ell = \sigma(W_\ell \mathbf{h}_{\ell-1} + \mathbf{b}_\ell), \quad f(\mathbf{x}) = W_{\mathrm{out}} \mathbf{h}_{L-1} + \mathbf{b}_{\mathrm{out}},
\end{equation}
where $\gamma$ is an optional input encoding, $\sigma$ is the nonlinearity, and hidden layers are indexed $\ell \in \{0, \ldots, L-1\}$. 
STRAINER~\citep{vyas2025learning} trains a single INR on $M$ signals $\{\mathbf{s}^{(j)}\}_{j=1}^M$ from a common domain by sharing the $L$ hidden layers as an encoder $f_\theta$ (with shared weights $\theta$) across all signals and equipping each signal with its own output projection $g_{\phi_j}$ (with per-signal weights $\phi_j$).
Writing $\Theta = (\theta, \{\phi_j\}_{j=1}^M)$ for the joint parameter set, training minimizes the reconstruction loss:
\begin{equation}
\Theta^\star = \arg\min_{\Theta} \sum_{j=1}^M \sum_{i=1}^N \lVert g_{\phi_j}(f_\theta(\mathbf{x}_i)) - \mathbf{s}^{(j)}_i \rVert_2^2.
\end{equation}

\paragraph{Layer-wise freezing protocol.}
The trained encoder $f_{\theta^\star}$, where $\theta^\star$ are the shared weights at the joint optimum, serves as initialization for new signals.
To isolate per-layer transferability, we sweep a freeze boundary $\tau \in \{0, \ldots, L-1\}$ over the hidden layers of the encoder. 
Layers $0, \ldots, \tau$ are held fixed at $\theta^\star$, while layers $\tau{+}1, \ldots, L-1$ are reinitialized alongside a fresh head $g_\phi$ and fit to a held-out target signal. 
% SIREN's first layer operates on a rank-2 coordinate input, while FFMLP operates on a high-dimensional Fourier embedding, thereby imposing distinct initial rank constraints on the learned weights.

\subsection{Locating the optimal freeze boundary via weight stable rank}
\label{sec:method-rank}

For a matrix $A$ with singular values $\sigma_1 \geq \sigma_2 \geq \dots$, the stable rank
\begin{equation}
\operatorname{sr}(A) = \frac{\lVert A \rVert_F^2}{\lVert A \rVert_2^2} = \frac{\sum_i \sigma_i^2}{\sigma_1^2}
\end{equation}
measures the effective dimensionality of the subspace spanned by $A$, weighted by the relative energy of its singular directions. 
Unlike algebraic rank, the stable rank is a continuous, noise-robust measure with $1 \leq \operatorname{sr}(A) \leq \operatorname{rank}(A)$~\citep{rudelson2005sampling}. 
Prior work has used the activation stable rank $\operatorname{sr}(H_\ell)$ as a measure of INR expressivity~\citep{ramasinghe2022beyond}. 
We track instead the weight stable rank $\operatorname{sr}(W_\ell^\star)$, an intrinsic property of the trained encoder. 
Each row of $W_\ell^\star$ is a direction in the input space of layer $\ell$ along which the layer reads its incoming activation, and $\operatorname{sr}(W_\ell^\star)$ measures how many such directions carry comparable energy. 
A high weight stable rank means the layer has come to span a broad subspace. A low weight stable rank means the layer has collapsed onto a few dominant directions~\citep{mcginnis2025optimizing}. 
This is distinct from the activation stable rank: a layer can span a broad weight subspace even if the current input distribution only excites part of it. 
For transferability the weight rank is the relevant measure, because the directions a layer can read are the directions a new signal can recruit, whether or not the cohort exercised them.

Sweeping the freeze boundary at test time (Sec.~\ref{sec:results-transfer}) reveals that the optimal freeze depth $\tau^\star$ aligns with the layer of peak weight stable rank in the cohort-trained encoder:
\begin{equation}
\label{eq:transfer-pred}
\tau^\star \in \arg\max_{\ell \in \{0, \ldots, L-1\}} \operatorname{sr}(W_\ell^\star).
\end{equation}
The alignment is consistent with a structural property of coordinate MLPs. 
The first trainable layer that operates on a high-rank input must embed the coordinate into a high-dimensional representation that separates nearby points, an operation requiring the layer's weight matrix to span a high-dimensional output. 
Peak weight stable rank therefore identifies the layer whose function inherently demands the most linear capacity, the \emph{coordinate embedding}. 
The two backbones place this embedding at different depths. 
FFMLP performs the embedding externally through its fixed Fourier projection, so $W_0^\star$ already operates on a high-rank input. 
SIREN has no fixed embedding and instead performs it internally through $W_1^\star$ acting on the post-sinusoid output of $W_0^\star$. 
Under cohort training, the embedding layer's weights come to encode cohort-wide coordinate structure that is reusable across signals. 
Freezing at $\tau^\star$ can be read as reusing this structure for new signals. 
The weight and activation rank profiles are correlated empirically, but they sometimes diverge in SIREN, where the weight rank aligns with $\tau^\star$ more accurately (Sec.~\ref{sec:results-transfer}).

\subsection{Sparse autoencoders as a probe of activation structure}
\label{sec:method-sae}

Sec.~\ref{sec:method-rank} identifies which layer transfers but not what that layer encodes, leaving the alignment between weight rank and $\tau^\star$ as an empirical observation without a representational account. 
To turn this observation into a mechanistic one, we need a basis in which each direction has an interpretable meaning, so that we can ask what the transferring layer's atoms actually represent and verify the answer through direct intervention. 
Neural features are rarely axis-aligned. 
Networks pack more features into non-orthogonal directions than they have neurons, a phenomenon known as \emph{superposition}~\citep{elhage2022toy,bricken2023monosemantic}. 
Linear methods like PCA cap the basis at the number of neurons and enforce orthogonality, collapsing this overcomplete geometry. 
Sparse autoencoders (SAEs) address this by learning an overcomplete dictionary with a sparse bottleneck, decomposing dense activations into a set of discrete \emph{atoms} that are directions in activation space the encoder uses to compose its outputs.

\paragraph{Architecture and training.}
For each layer, a separate SAE operates on the hidden activations $\mathbf{h} \in \mathbb{R}^d$, since the dictionary at each depth captures different structure. We follow the TopK architecture~\citep{gao2024scaling}, which controls the sparsity level $k=32$ explicitly and leaves feature magnitudes unbiased, unlike $\ell_1$ penalties~\citep{makhzani2013k}. The encoder maps mean-centered activations to a dictionary of $n = 4096$ atoms:
\begin{equation}
\mathbf{z} = \mathrm{TopK}\!\left(\mathrm{ReLU}\!\left(W_{\mathrm{enc}} (\mathbf{h} - \mathbf{b}_{\mathrm{pre}}) + \mathbf{b}_{\mathrm{enc}}\right), k\right), \quad W_{\mathrm{enc}} \in \mathbb{R}^{n \times d},
\end{equation}
where $\mathbf{b}_{\mathrm{pre}}$ centers the activation before encoding and is added back after decoding. The decoder reconstructs the activation as a linear combination of atoms:
\begin{equation}
\hat{\mathbf{h}} = W_{\mathrm{dec}}^\top \mathbf{z} + \mathbf{b}_{\mathrm{pre}}, \quad W_{\mathrm{dec}} \in \mathbb{R}^{n \times d}.
\end{equation}
We constrain the rows of $W_{\mathrm{dec}}$ (the atoms) to unit norm throughout training. This prevents magnitude shrinkage and ensures $z_a$ directly reflects the strength of atom $a$ in the input.

\paragraph{Atom maps and ablations.}
We evaluate SAE atoms via two diagnostics. The \emph{atom-activation map} $z_a(\mathbf{x})$ visualizes the spatial firing pattern of atom $a$ across the input domain. The \emph{single-atom ablation} measures atom $a$'s causal contribution by replacing the activation $H_\ell$ with $H_\ell - z_a w_a^\top$, where $w_a$ is the $a$-th row of $W_{\mathrm{dec}}$, and measuring the pixel-space difference $\Delta_a(\mathbf{x}) = f_\theta(\mathbf{x}) - f_\theta^{-a}(\mathbf{x})$. Together, these reveal where an atom fires and what its removal affects (Sec.~\ref{sec:results-dictionaries}).
% ----------------------------------------------------------------------
\section{Experiments}
\label{sec:results}

\paragraph{Architectures and training}
We adopt SIREN~\citep{sitzmann2020implicit} and FFMLP~\citep{tancik2020fourier}, both with $5$ hidden layers of width $d = 256$, and consider two training regimes per backbone. \emph{Single-signal} INRs fit one image from random init for $500$ iterations. \emph{Cohort-trained} INRs follow STRAINER~\citep{vyas2025learning}, jointly fitting $M = 10$ signals through a shared $5$-layer encoder and per-signal heads for $5000$ iterations to cohort-mean PSNR $\approx 30$ dB. At test time, the cohort encoder is reused as initialization and refit with the same $500$-iteration budget as single-signal. The four conditions used throughout are single-signal SIREN-FFMLP, and cohort SIREN-FFMLP. Full hyperparameters in Appendix~\ref{sec:appendix-strainer-comparison}.

\paragraph{Datasets.}
CelebA-HQ~\citep{karras2017progressive} at $178 \times 218$ provides aligned face photographs. OASIS-MRI~\citep{Marcus2007OpenAS} at $160 \times 192$ provides template-aligned axial brain MRI slices. Kodak~\citep{kodak1993kodak} at $768 \times 512$ provides high-resolution natural photographs. We keep each image at its native resolution. For each dataset we randomly select $10$ images for cohort training and $100$ test images for evaluation, with Kodak's $24$-image total restricting us to $14$ test images.

\subsection{Localizing transferable representations in cohort INRs}
\label{sec:results-transfer}

\begin{figure*}[t]
    \centering
    \includegraphics[width=\textwidth]{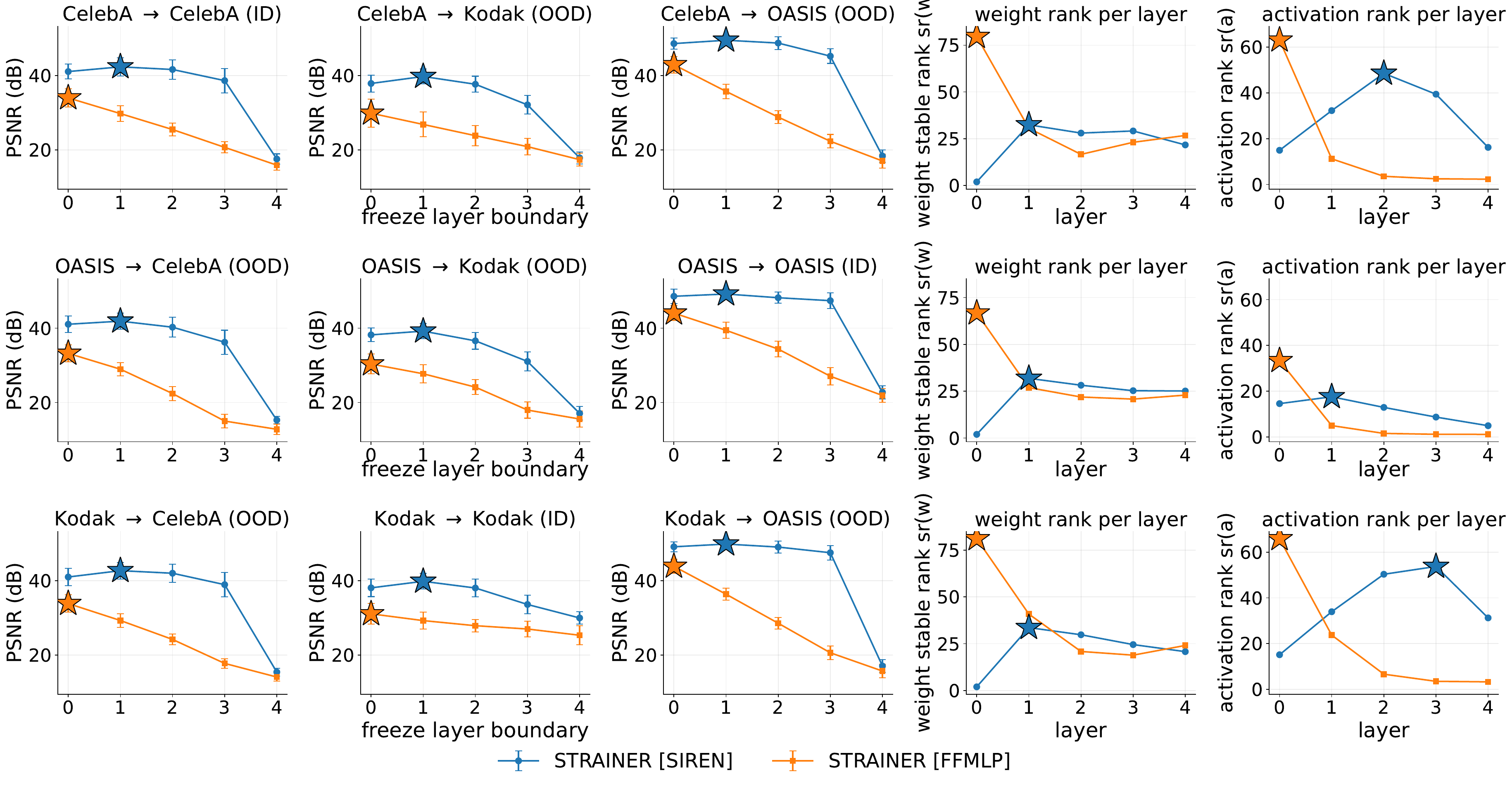}
    \caption{
        \textbf{Layer-wise transfer sweep for cohort-trained SIREN and FFMLP.}
        Three cohort sources (CelebA, OASIS, Kodak) on three target distributions each. 
        Columns 1-3: PSNR (dB) vs.\ freeze boundary $\tau$, mean $\pm$ std across fitting seeds. 
        Columns 4-5: weight stable rank $\operatorname{sr}(W_\ell)$ and activation stable rank $\operatorname{sr}(H_\ell)$ per layer of the cohort-trained encoder. 
        Stars mark the argmax in each panel. 
        The layer of peak weight stable rank aligns with the empirical $\tau^\star$ in all six (source, architecture) configurations.
    }
    \label{fig:rank-sweep}
\end{figure*}

\paragraph{Setup.}
We apply the freeze sweep only to cohort-trained INRs.
For each architecture, source, and target dataset, we sweep the freeze boundary $\tau \in \{0, 1, \ldots, L-1\}$ and fit each configuration to each test image using three reinitialization seeds. 
We report PSNR (mean $\pm$ standard deviation across test images and seeds) at the end of the test-time fit (Fig.~\ref{fig:rank-sweep}, Tab.~\ref{tab:ours-vs-strainer}). 
Per-dataset SSIM and LPIPS~\citep{zhang2018unreasonable} can be found in Tabs.~\ref{tab:strainer-celeba}, \ref{tab:strainer-kodak}, and~\ref{tab:strainer-oasis} of the Appendix.

\paragraph{Results.}
Across all three cohorts and all three target distributions, transferable representations localize to a single layer in the encoder. 
The optimal freeze depth is invariant within each architecture: $\tau^\star = 1$ for SIREN and $\tau^\star = 0$ for FFMLP (Fig.~\ref{fig:rank-sweep}). 
At this depth, the freeze protocol matches STRAINER's initialization recipe in most cohort-target combinations and substantially exceeds it on Kodak, where it gains $8$-$10$ dB over the full recipe (Tab.~\ref{tab:ours-vs-strainer}). 
Two distinct effects produce this gap. 
For Kodak~$\to$~Kodak, both protocols start from a Kodak-trained encoder, but the $500$-iteration test-time budget is not enough for the no-freeze baseline to converge across all encoder and decoder layers at Kodak's high resolution. 
For CelebA/OASIS~$\to$~Kodak, the cohort encoder is trained at the lower source resolution and applied to Kodak's higher coordinate density, so the no-freeze baseline must additionally rebuild every layer from a mismatched initialization within the same budget. 
Freezing the lower layers reduces the parameters that update at test time, addressing both effects at once.

\begin{table}[t]
\centering
\small
\setlength{\tabcolsep}{4pt}
\caption{
\textbf{Test-time PSNR} (dB, mean $\pm$ std).
STRAINER's (no-freeze) recipe vs.\ our freeze protocol at $\tau^\star$ ($\tau{=}1$ for SIREN, $\tau{=}0$ for FFMLP). 
The $W_\ell / H_\ell$ column indicates whether the layer of peak weight/activation stable rank in the cohort-trained encoder coincides with the empirical $\tau^\star$ for that source. 
The freeze protocol matches STRAINER in most combinations and substantially exceeds it on Kodak, with mean gains of $+3.1$ dB for SIREN and $+1.3$ dB for FFMLP across all combinations.
}
\label{tab:ours-vs-strainer}
\begin{tabular}{l c cc c cc}
\toprule
 & \multicolumn{3}{c}{\textbf{STRAINER [SIREN]}} & \multicolumn{3}{c}{\textbf{STRAINER [FFMLP]}} \\
\cmidrule(lr){2-4} \cmidrule(lr){5-7}
Source $\to$ Target & $W_\ell / H_\ell$ & No freeze & $\tau=1$ & $W_\ell / H_\ell$ & No freeze & $\tau=0$ \\
\midrule
CelebA $\to$ CelebA & \multirow{3}{*}{\cmark / \xmark} & \textbf{43.15}{ $\pm$2.26} & 42.36{ $\pm$2.47} & \multirow{3}{*}{\cmark / \cmark} & 33.20{ $\pm$2.42} & \textbf{34.01}{ $\pm$2.40} \\
CelebA $\to$ Kodak  & & 30.93{ $\pm$2.65} & \textbf{39.77}{ $\pm$1.99} & & 28.43{ $\pm$2.54} & \textbf{29.89}{ $\pm$3.78} \\
CelebA $\to$ OASIS  & & 49.28{ $\pm$1.30} & \textbf{49.52}{ $\pm$1.42} & & 39.61{ $\pm$1.57} & \textbf{42.93}{ $\pm$2.21} \\
\midrule
Kodak $\to$ CelebA & \multirow{3}{*}{\cmark / \xmark} & \textbf{43.41}{ $\pm$2.08} & 42.67{ $\pm$2.21} & \multirow{3}{*}{\cmark / \cmark} & 33.61{ $\pm$2.23} & \textbf{33.83}{ $\pm$2.35} \\
Kodak $\to$ Kodak  & & 30.02{ $\pm$2.63} & \textbf{39.81}{ $\pm$2.11} & & 28.76{ $\pm$2.62} & \textbf{31.06}{ $\pm$2.78} \\
Kodak $\to$ OASIS  & & 49.78{ $\pm$1.00} & \textbf{49.83}{ $\pm$1.42} & & 40.82{ $\pm$1.49} & \textbf{43.84}{ $\pm$1.87} \\
\midrule
OASIS $\to$ CelebA & \multirow{3}{*}{\cmark / \cmark} & 40.12{ $\pm$3.34} & \textbf{41.87}{ $\pm$2.15} & \multirow{3}{*}{\cmark / \cmark} & \textbf{34.71}{ $\pm$3.15} & 33.21{ $\pm$2.30} \\
OASIS $\to$ Kodak  & & 30.56{ $\pm$2.68} & \textbf{39.19}{ $\pm$2.07} & & 28.66{ $\pm$2.57} & \textbf{30.35}{ $\pm$2.67} \\
OASIS $\to$ OASIS & & 48.98{ $\pm$2.02} & \textbf{49.17}{ $\pm$1.39} & & \textbf{44.11}{ $\pm$2.21} & 44.04{ $\pm$2.01} \\
\midrule
\textit{Mean across combinations} & & 40.69 & \textbf{43.80} & & 34.66 & \textbf{35.91} \\
\bottomrule
\end{tabular}
\end{table}

Weight stable rank aligns with $\tau^\star$ in all $6$ (source, architecture) configurations (Tab.~\ref{tab:ours-vs-strainer}). 
Activation stable rank agrees in only $4$ of $6$, with $\operatorname{sr}(H_\ell)$ peaking at deeper layers than $\tau^\star$ for CelebA-source and Kodak-source SIREN. 
The divergence is consistent with the argument in Sec.~\ref{sec:method-rank}: $\operatorname{sr}(H_\ell)$ depends on the cohort that exercised the layer, while $\operatorname{sr}(W_\ell)$ does not. 
The architectural offset $\tau^\star = 1$ vs.\ $0$ matches the picture in Sec.~\ref{sec:method-rank}: FFMLP's fixed Fourier projection performs the coordinate embedding before any trainable layer, so $W_0$ already operates on a high-rank input, while SIREN performs the embedding internally through $W_1$.

% The PSNR curve shapes already hint at the dictionary asymmetry developed in Sec.~\ref{sec:results-dictionaries}. 
% SIREN plateaus through $\tau \in \{0, 1, 2\}$ within $\sim 1$ dB of the peak before collapsing at $\tau = 4$, suggesting transferable structure is present across the lower layers, with $\tau^\star = 1$ as the empirical best. 
% FFMLP peaks sharply at $\tau = 0$ and drops by $4$-$8$ dB per additional frozen layer, suggesting transferable structure is concentrated at $\tau = 0$ and that the layers above it encode something the test-time fitter would prefer to overwrite. 
% The dictionary analysis in Sec.~\ref{sec:results-dictionaries} traces this difference to what each architecture's atoms encode.

\subsection{Atom dictionaries split by architecture, not training regime}
\label{sec:results-dictionaries}

\paragraph{Setup.}
We fit a separate SAE to the post-activations of each hidden layer of the trained INR encoder, which remains fixed. 
The dictionary is overcomplete with $n = 4096$ atoms ($16\times$ the INR hidden width $d = 256$) to accommodate superposition~\citep{elhage2022toy}, and TopK with $k = 32$ balances reconstruction fidelity against feature specialization. 
Each SAE is trained for $10$k steps with Adam at learning rate $10^{-4}$ and decoder atoms constrained to unit norm. 
We characterize each atom by two statistics: its \emph{active-pixel fraction} (coordinates where $z_a > 0.01$, a small threshold above the noise floor) and the layer-wise \emph{dead-atom rate} (atoms that never fire), measuring spatial spread and effective dictionary capacity. 
Ablations on $n$ and $k$ and SAE diagnostics are in \autoref{sec:appendix-sae}.

\begin{figure*}[t]
    \centering
    \includegraphics[width=0.98\textwidth]{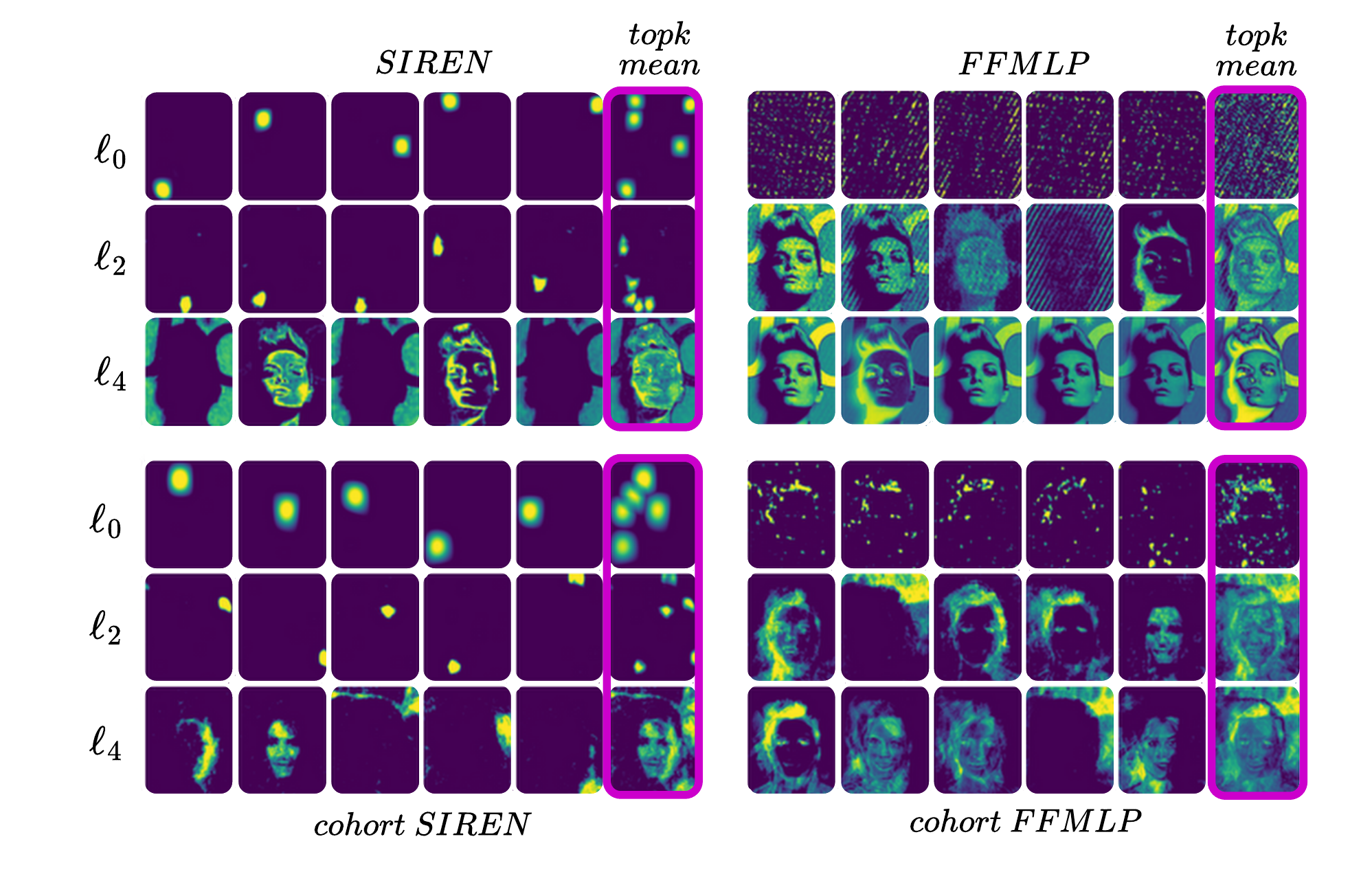}
    \caption{
    \textbf{SAE atom dictionaries split by architecture, not training regime.}
    Top-$5$ atoms by mean magnitude at layers $\ell_0$, $\ell_2$, $\ell_4$ for cohort-trained SIREN and FFMLP. 
    The sixth column (magenta border) shows the average of the top-$k$ atoms at that depth. 
    SIREN atoms are spatially localized blobs that tile the coordinate plane at every depth, content-independent positional primitives learned over the cohort. 
    FFMLP atoms trace cohort image content from $\ell_2$ onward, with the top-$k$ mean column of Fig.~\ref{fig:rank-sweep} resolves into the cohort face that this concentration implies. 
    At $\ell_0$, FFMLP shows Fourier-residue speckle, consistent with the neuron-level Fourier signatures reported by~\citep{padmanabhan2024explaining}.
}
    \label{fig:atom-gallery}
\end{figure*}

\paragraph{Results.}
The two architectures recover qualitatively different dictionaries despite training on the same cohort to comparable PSNR. 
SIREN's atoms encode \emph{where} a coordinate sits in the image plane: each atom fires in a confined spatial region, and the dictionary as a whole tiles the coordinate domain (Fig.~\ref{fig:atom-gallery}). 
FFMLP's behavior splits by depth. 
At $\ell_0$, atoms show diagonal speckle patterns that reflect the fixed Fourier basis the layer reads, consistent with the neuron-level Fourier signatures reported by~\citep{padmanabhan2024explaining}. 
From $\ell_2$ onward, atoms instead encode \emph{what} signal content lives at each coordinate: individual atoms trace the contours of memorized cohort signals (faces in CelebA, brain anatomy in OASIS, see also \autoref{sec:appendix-galleries}). 
The split tracks architecture, not training regime: a single SIREN trained on one image still tiles the coordinate plane at $\ell_0$-$\ell_2$, and a single FFMLP still collapses into image-shaped atoms from $\ell_2$ onward.

\begin{figure}[t]
    \centering
    \includegraphics[width=\textwidth]{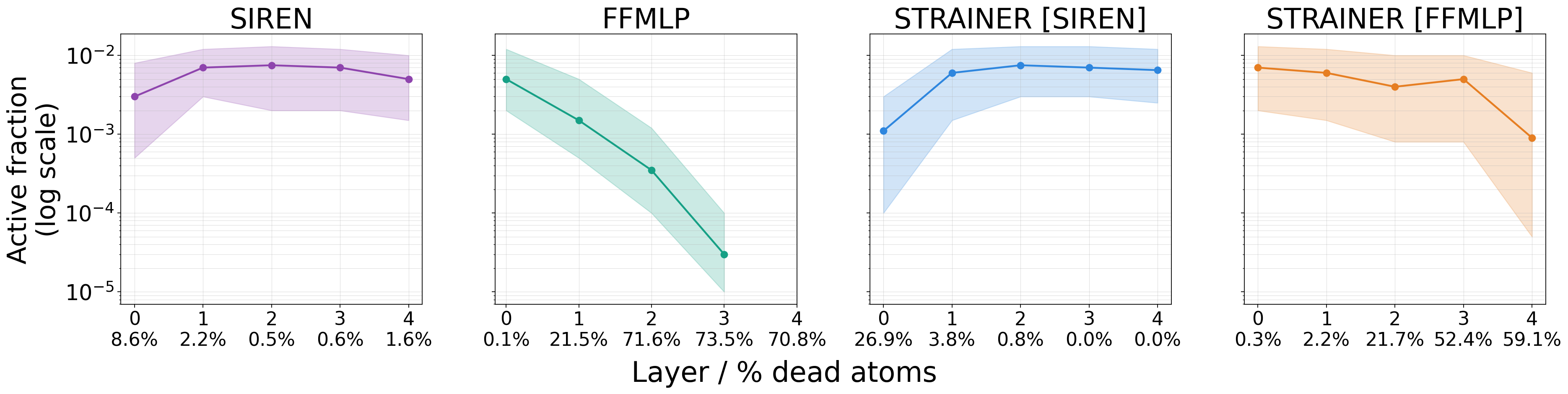}
\caption{
    \textbf{Atom-level concentration and dictionary capacity across depth.}
    Active-pixel fraction (share of input coordinates at which an atom fires) per layer. Median over alive atoms, shaded IQR ($25^{\text{th}}$-$75^{\text{th}}$ percentile). Below each layer index: percentage of dead atoms. 
    Cohort SIREN keeps its dictionary alive (dead atoms $\leq 4\%$ from $\ell_1$ onward) with each atom firing on $\sim 1\%$ of pixels. Cohort FFMLP shrinks its usable dictionary (dead atoms $0.3\% \to 59.1\%$ across depth) and concentrates the survivors onto $30$-$50\%$ of pixels each. Single-signal variants follow the same architectural split.
}
    \label{fig:concentration}
\end{figure}

Fig.~\ref{fig:concentration} makes the split quantitative. Cohort SIREN keeps its dictionary alive with sparse atoms, the signature of a tiling positional code. Cohort FFMLP shrinks its usable dictionary and concentrates the survivors onto a large share of pixels each, the signature of memorization, with the top-$k$ mean column of Fig.~\ref{fig:atom-gallery} resolving into the cohort face this implies.

The asymmetry follows from where each architecture builds its coordinate embedding. 
FFMLP receives a positional code from its fixed Fourier projection, so the trainable stack inherits a usable embedding at $W_0$ and the layers downstream have spare capacity that cohort signals fill with memorized content. 
SIREN has no fixed projection, so the stack must build the coordinate code itself, and cohort training shapes this construction toward a content-independent tiling. 
This split also explains the freeze-sweep curves of Sec.~\ref{sec:results-transfer}: SIREN's plateau across $\tau \in \{0, 1, 2\}$ with $\tau = 1$ to be the empirical optimum before collapsing at $\tau = 4$, reflects reusable positional structure the test-time fitter inherits intact, while FFMLP's $4$-$8$ dB drop per additional frozen layer reflects memorized content the fitter must work against.
This is also observed in single atom ablations (Sec.~\ref{sec:results-interventions}).

\subsection{Single-atom ablations confirm causal use of the dictionaries}
\label{sec:results-interventions}

\begin{figure}[t]
    \centering
    \includegraphics[width=0.85\textwidth]{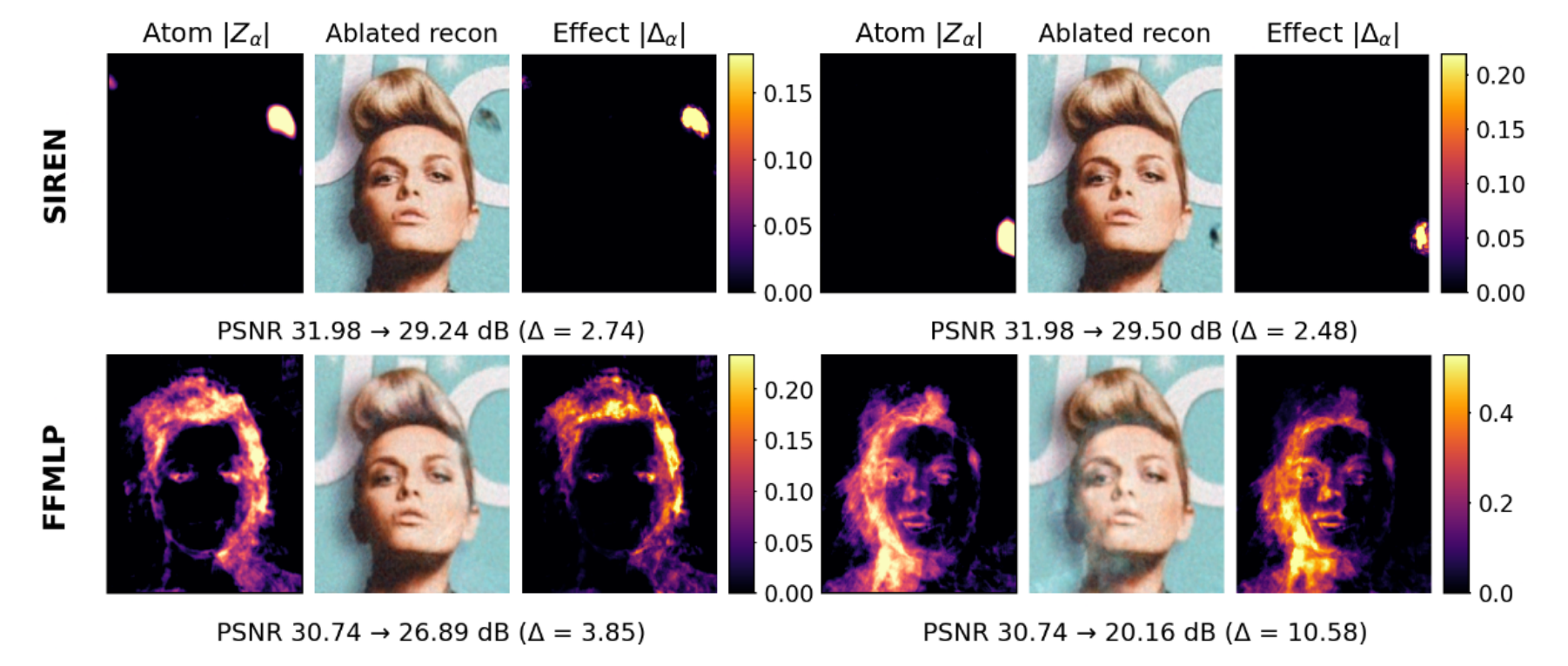}
\caption{
    \textbf{SIREN damage stays local, FFMLP damage spreads globally.}
    Single-atom ablations at $\ell_2$ on a CelebA test image, two example atoms per architecture from cohort-trained INRs. 
    For each atom: \emph{Atom $|Z_a|$} shows where the atom fires, \emph{Ablated recon} the network output after zeroing the atom, and \emph{Effect $|\Delta_a|$} the per-pixel change in reconstruction. 
    PSNR before $\to$ after ablation below each example. 
    A single FFMLP atom out of $4096$ drops image-wide PSNR by up to $10.58$ dB, while SIREN ablations stay confined to the firing region. 
    All-layer ablations in \autoref{sec:appendix-ablations}.
}
    \label{fig:atom-ablation}
\end{figure}

\paragraph{Setup.}
We test whether the network uses its dictionary in the way Sec.~\ref{sec:results-dictionaries} suggests. 
If SIREN's atoms encode reusable spatial regions, ablating one should affect only where it fires.
Similarly, if FFMLP's deep atoms carry image-scale content, ablating one should affect the entire reconstruction. We ablate the top-$2$ atoms by mean magnitude choosing an intermediate layer, i.e., $\ell_2$, and report the resulting PSNR drop. 
All-layer ablations in \autoref{sec:appendix-ablations}.
\paragraph{Results.}
SIREN ablations produce \emph{spatially localized} effects: the per-pixel reconstruction error is supported on the small region where the atom fires, and PSNR drops by $2$-$3$ dB concentrated in that region (Fig.~\ref{fig:atom-ablation}). 
FFMLP ablations produce \emph{globally distributed} effects: the error spreads across the entire image, often well beyond the atom's nominal firing region. 
A single FFMLP atom out of $4096$ can drop image-wide PSNR by up to $10.58$ dB, evidence that the network depends on each surviving deep atom for global signal reconstruction.

The asymmetry traces to how each architecture distributes signal across its dictionary at the ablation layer (full statistics in \autoref{sec:appendix-sae}). 
Cohort SIREN spreads reconstruction across many atoms of similar magnitude ($\sim 0.05$ each), each firing on $\sim 2\%$ of pixels. 
Removing one therefore costs only the region it covers, and other atoms continue to support the rest of the image. 
Cohort FFMLP concentrates reconstruction onto a few high-magnitude atoms (top atom $\sim 0.35$, roughly $7\times$ the next tier), each firing on $> 40\%$ of pixels and carrying image-scale content. 
Removing one of these atoms cannot be compensated by the others, and the entire reconstruction degrades. 
The magnitude and fire-rate profiles jointly imply this dependence, and the ablations confirm it.

\section{Discussion and Conclusion}

\paragraph{Limitations.}
While this work rests on two backbones, SIREN and FFMLP, the mechanism we identify has not been tested on other prominent INR architectures such as Gaussian-activation networks~\citep{ramasinghe2022beyond}, FINER~\citep{liu2024finer}, WIRE~\citep{saragadam2023wire}, or hash grids~\citep{muller2022instant}, and extending the analysis to these is the natural next step. 
We also fix INR depth and width and use one SAE configuration $(n, k) = (4096, 32)$ across all experiments. Our SAE ablations in Appendix~\ref{sec:appendix-sae} suggest the trends we report are stable around this configuration, but how the SAE decomposition itself changes with INR depth and width remains open.
Finally, although the SAE recovers atoms with recognizable spatial firing patterns, we do not characterize what individual atoms encode semantically (color, edges, identity, anatomical structure), nor do we explore steering reconstructions by intervening on atoms beyond the single-atom ablations we report. Both are left to future work.

\paragraph{Conclusion.}
% This work characterizes what cohort-trained INRs encode and where to freeze them, providing a weight-rank diagnostic for transfer-depth selection and an SAE decomposition that turns INR activations into an inspectable basis. Both contributions reduce the per-signal cost of fitting INRs, which can broaden the applicability of INRs to data-constrained settings such as patient-specific medical imaging, personalized signal representation, and real-time domains like video streaming and robotics. The diagnostic is a one-shot computation on the encoder, removing a hyperparameter search that would otherwise scale with the cost of test-time fitting. The SAE decomposition opens a path toward mechanistic interpretability of coordinate-based networks, which has been largely studied through spectral and kernel-theoretic lenses rather than direct activation decomposition. Our methods analyze and improve INRs as a general signal-representation tool, and the same techniques apply regardless of the underlying intent of downstream applications.
We set out to investigate whether reusing parts of a cohort-trained INR by freezing them at test time makes sense, and we found that it does, but only at a specific architecture-dependent depth. 
Across SIREN and FFMLP, the empirical optimal depth coincides with the layer of highest weight stable rank, indicating that the shared layer at this depth carries structure learned over the cohort that is reusable across signals.
The rank profiles recast the architectural split in encoder-decoder terms. FFMLP comes with a fixed encoder (the Fourier projection) and learns only the decoder, while SIREN learns both, with the rank lift inside $W_1$ marking where its encoder ends and its decoder begins.
SAE visualizations, atom-level statistics, and single-atom ablations together show that SIREN's atoms are spatially localized and tile the coordinate plane, while FFMLP's deep atoms are spatially diffuse and encode memorized cohort content.
Our protocol is orthogonal to the cohort recipe since it is a post-hoc analysis of the trained encoder, so the same diagnostic can be applied to STRAINER~\citep{vyas2025learning}, SNP~\citep{vyas2026surprising}, and other cohort approaches alike. 
Reducing the per-signal cost of fitting INRs by reusing structure learned from related signals would significantly broaden the applicability of INRs, and our findings suggest a concrete handle on which structure can be reused. 
Beyond the immediate setting, we hope the weight-rank diagnostic and the SAE decomposition open a path toward characterizing what INRs encode and toward architectures designed for generalization rather than memorization. 
Both tools operate on trained INR weights and activations directly, neutral to the application that produced the INR.

\bibliographystyle{plainnat}
\bibliography{bibliography}

% -----------------------------------------------------------------------
\clearpage
\appendix
\label{sec:Appendix}

\section{SAE supplementary material}
\label{sec:appendix-sae}

\subsection{SAE hyperparameters}
\label{sec:appendix-sae-hyp}
\begin{figure}[!htbp]
  \centering
  \includegraphics[width=\textwidth]{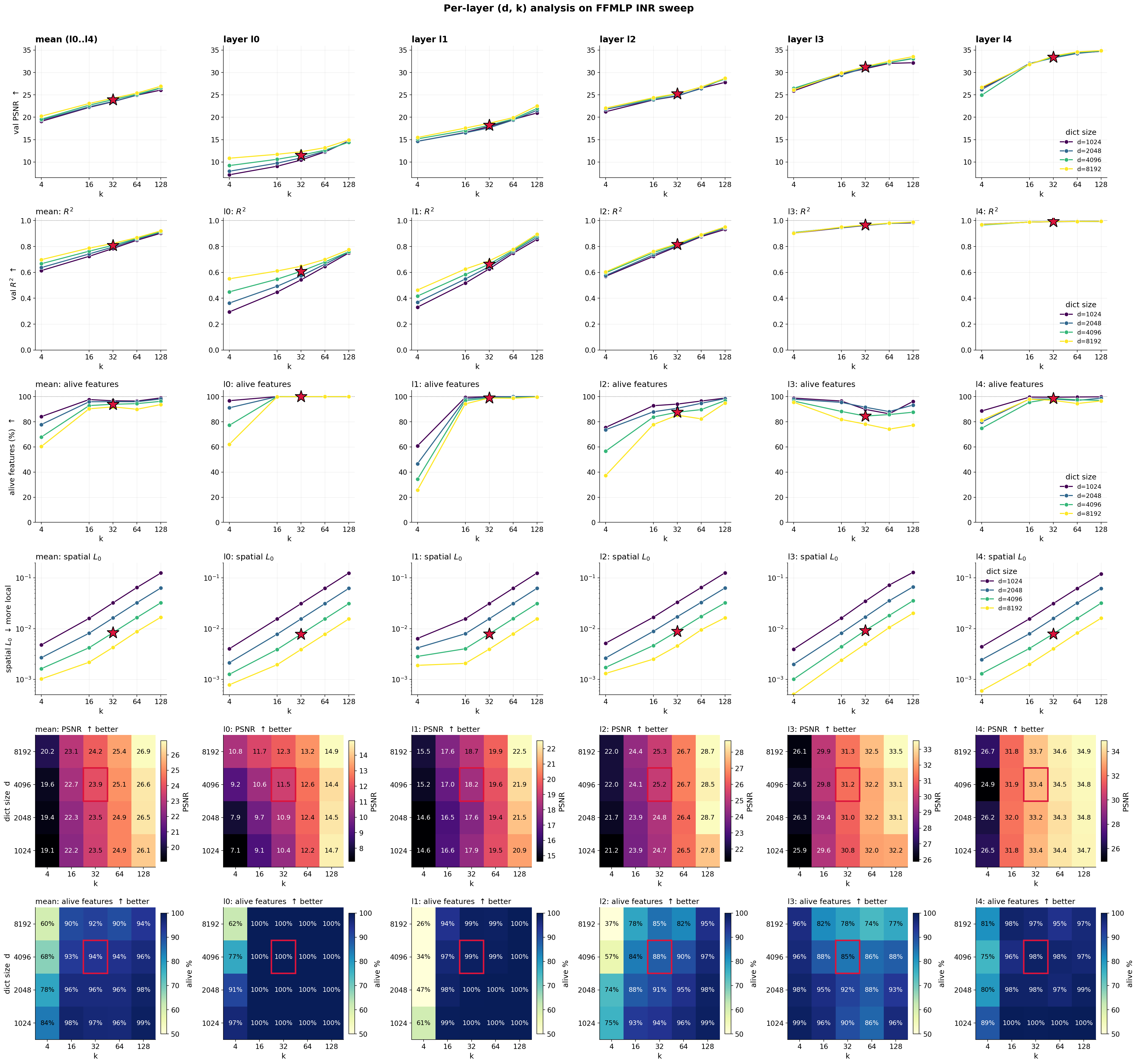}
  \caption{
  \textbf{Per-layer $(n, k)$ analysis on the FFMLP INR sweep.}
  Columns: across-layer mean (leftmost) and individual layers $\ell_0$-$\ell_4$. Rows: (i) reconstruction PSNR, (ii) $R^2$, (iii) alive features (\%), (iv) spatial $L_0$ (lower = more local), (v) PSNR heatmap over $(n, k)$, (vi) alive features heatmap. Curves are colored by $n$. The chosen operating point $(n, k) = (4096, 32)$ is marked by red stars on the line panels and red rectangles on the heatmaps.
  }
  \label{fig:dk_ffmlp}
\end{figure}

\begin{figure}[!htbp]
  \centering
  \includegraphics[width=\textwidth]{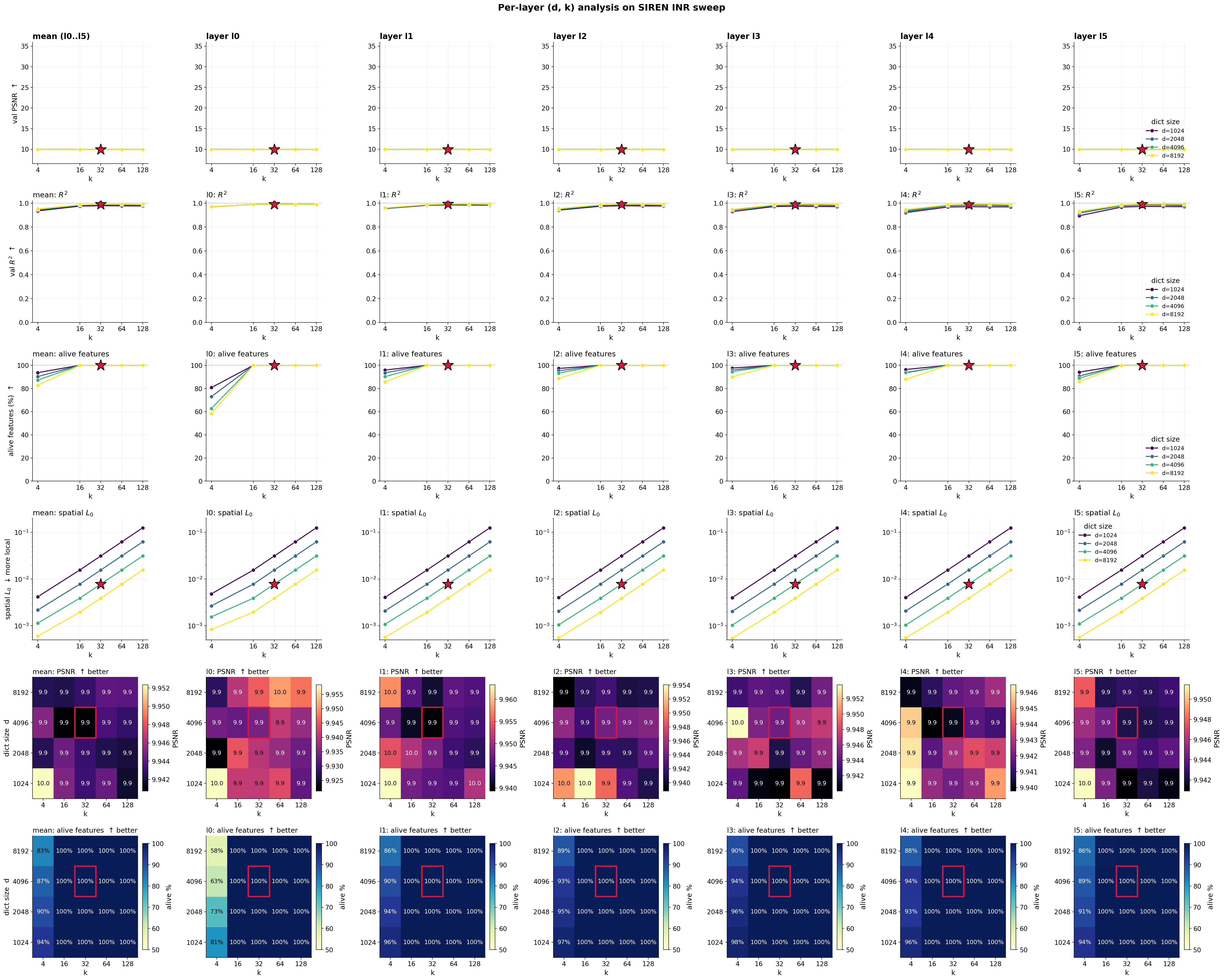}
  \caption{
  \textbf{Per-layer $(n, k)$ analysis on the SIREN INR sweep}, in the same layout as Fig.~\ref{fig:dk_ffmlp}. The base SIREN INR underfits at this configuration, so the downstream-image PSNR row (i) and heatmap (v) are essentially constant across the grid and should not be read as informative. The remaining rows ($R^2$, alive \%, spatial $L_0$) measure the SAE's reconstruction of activations directly and remain meaningful: $R^2$ is uniformly high, alive \% saturates near $100\%$ except at $k = 4$, and spatial $L_0$ shows the same strict monotone decrease in $n$ observed for FFMLP. The chosen operating point is highlighted as before.
  }
  \label{fig:dk_siren}
\end{figure}

We sweep dictionary size $n \in \{1024, 2048, 4096, 8192\}$ and TopK sparsity $k \in \{4, 16, 32, 64, 128\}$ for each backbone, evaluating four diagnostics per $(n, k)$ cell. Reconstruction fidelity is reported as PSNR of the image obtained by substituting SAE-reconstructed activations into the INR, and as per-layer $R^2 = 1 - \mathrm{MSE}/\mathrm{Var}$ (bounded in $[0,1]$, with saturation more visible than in PSNR). Dictionary utilization (alive \%) is the fraction of atoms that fire at least once, the standard diagnostic against dead-feature collapse. Spatial $L_0$ is the mean fraction of coordinates at which an atom fires, with lower values corresponding to more localized atoms. We choose $(n, k) = (4096, 32)$ as the point that scores well on every axis without compromising any of them: reconstruction saturates well before $n = 4096$ (doubling $n$ beyond $2048$ yields under $0.5$ dB mean PSNR on FFMLP and negligible $R^2$ gain on both sweeps), $k = 32$ stays clear of the dead-feature regime that emerges at $k \leq 16$ when $n \geq 4096$, and dictionary utilization remains $\geq 85\%$ at every layer. The decisive criterion is spatial $L_0$, which decreases strictly with $n$ at every (layer, $k$) cell: a larger dictionary yields more localized atoms. Combined with the reconstruction saturation, $n = 4096$ buys locality (and a larger pool of usable feature slots) at no fidelity cost. The same ordering holds across both backbones (Fig.~\ref{fig:dk_ffmlp}, Fig.~\ref{fig:dk_siren}), so the choice is not architecture-specific.

\subsection{SAE health diagnostics}
\label{sec:appendix-sae-health}

\begin{figure}[h]
    \centering
    \includegraphics[width=0.7\linewidth]{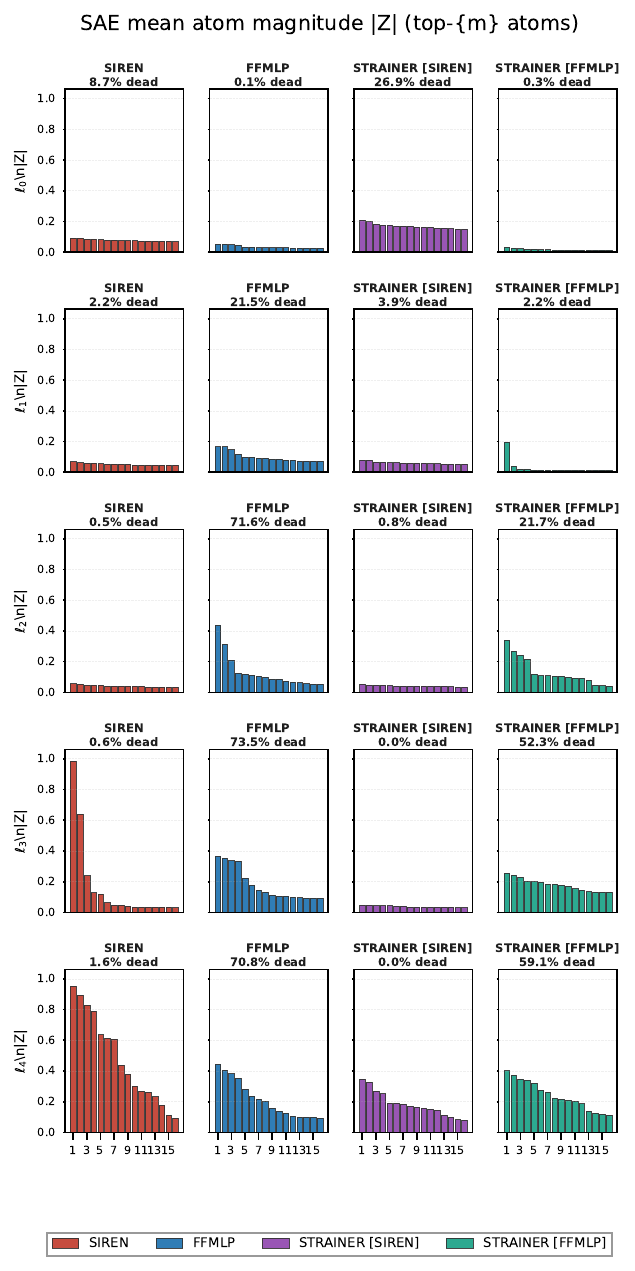}
    \caption{
        \textbf{SAE mean atom magnitude $|Z|$ for the top-$16$ atoms at each layer of each cell.}
        Panel titles report the per-cell dead-atom fraction. Single-signal SIREN and cohort SIREN show flat magnitude profiles at every depth (top atom $\sim 0.05$), with single-signal SIREN developing a steep dominance at $\ell_3$ and $\ell_4$ that cohort training counteracts. Single-signal FFMLP and cohort FFMLP show pronounced magnitude concentration from $\ell_1$ onward, with the top atom carrying $7$-$10\times$ the magnitude of the next tier from $\ell_2$ to $\ell_4$.
    }
    \label{fig:appendix-atom-magnitude}
\end{figure}

\begin{figure}[h]
    \centering
    \includegraphics[width=0.7\linewidth]{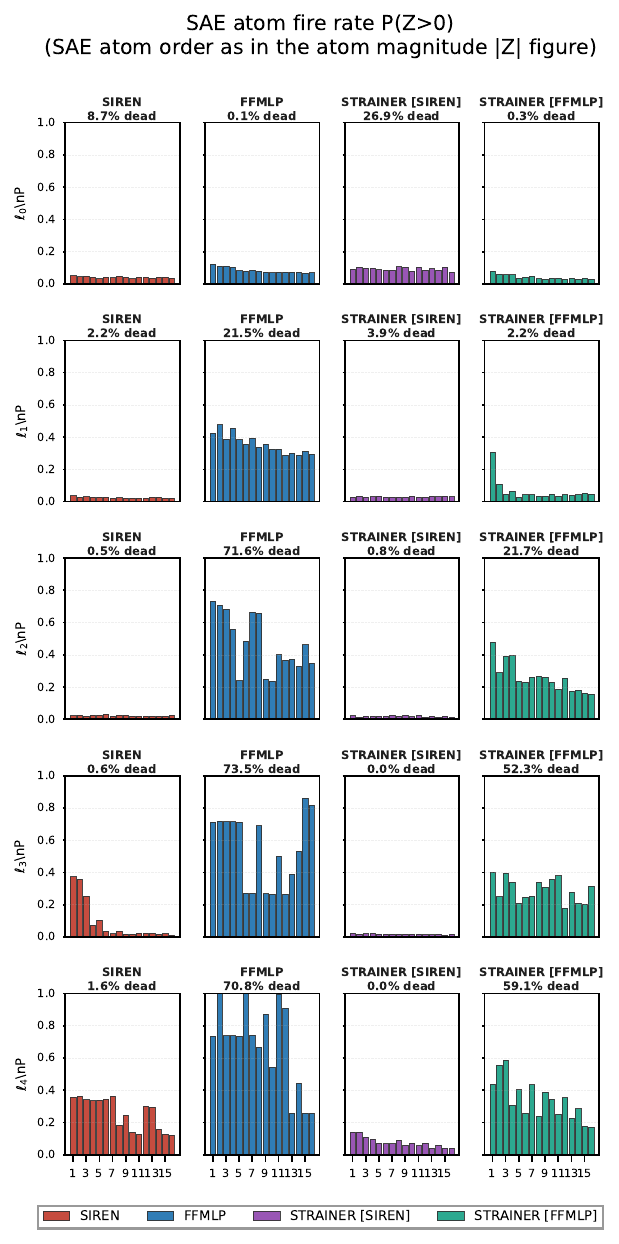}
    \caption{
        \textbf{SAE per-pixel firing rate $P(Z > 0)$ for the top-$16$ atoms at each layer of each cell.}
        Atoms ordered as in Fig.~\ref{fig:appendix-atom-magnitude}. Panel titles report the per-cell dead-atom fraction. SIREN cells maintain sparse firing at all depths ($P \lesssim 0.05$), consistent with spatially localized atoms. Single-signal FFMLP and cohort FFMLP show firing rates rising sharply with depth, with the surviving top atoms reaching $P \approx 0.7$-$0.9$ at $\ell_3$ and $\ell_4$: each surviving atom fires across most of the image.
    }
    \label{fig:appendix-fire-rate}
\end{figure}

\begin{figure}[h]
    \centering
    \includegraphics[width=\linewidth]{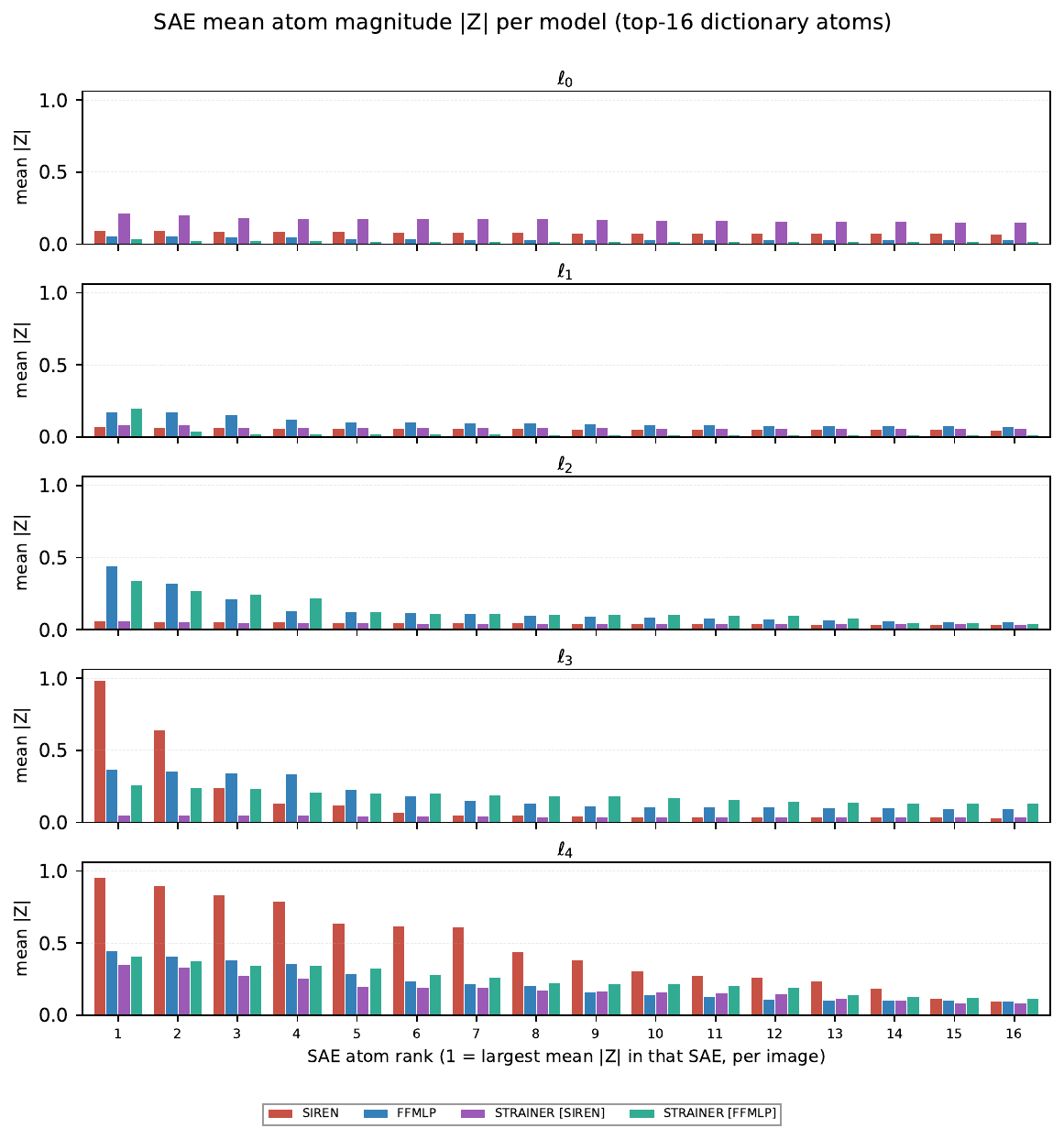}
    \caption{
        \textbf{Mean atom magnitude $|z_a|$ for the top-$16$ atoms across the four cells, overlaid per layer.} Each layer panel shows the magnitude profile for single-signal SIREN, single-signal FFMLP, cohort SIREN, and cohort FFMLP in matched atom-rank ordering.
    }
    \label{fig:appendix-magnitude-comparison}
\end{figure}

\begin{figure}[h]
    \centering
    \includegraphics[width=\linewidth]{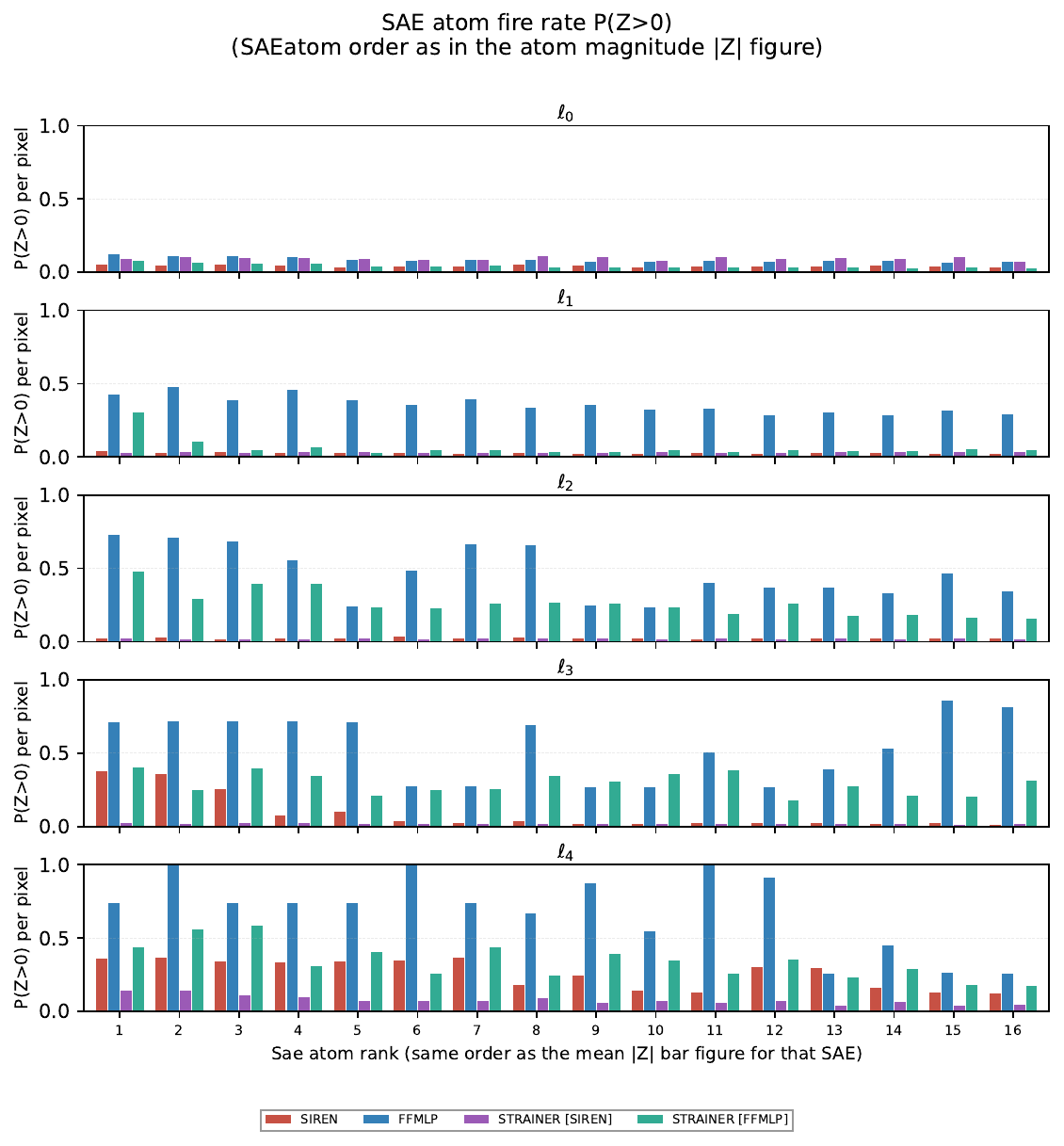}
    \caption{
        \textbf{Per-pixel firing rate $P(z_a > 0)$ for the top-$16$ atoms across the four cells, overlaid per layer.}
        Each layer panel shows the firing-rate profile for single-signal SIREN, single-signal FFMLP, cohort SIREN, and cohort FFMLP in matched atom-rank ordering. SIREN cells (red, purple) maintain low firing rates ($P \lesssim 0.05$) at every layer except layer $4$ of single-signal SIREN, consistent with sparse, spatially localized atom activation. FFMLP cells (blue, green) exhibit firing rates rising to $0.7$-$0.9$ in deeper layers, with the surviving atoms each firing across most of the image. The dichotomy mirrors the magnitude comparison in Fig.~\ref{fig:appendix-magnitude-comparison}: SIREN distributes representation across many sparsely-firing atoms, FFMLP concentrates representation in a few densely-firing atoms.
    }
    \label{fig:appendix-fire-rate-comparison}
\end{figure}

We characterize SAE health across all (cell, layer) pairs through two complementary diagnostics: per-pixel firing rate $P(z_a > 0)$ and mean atom magnitude $|z_a|$, computed for the top-$16$ atoms in each SAE. 
Figs.~\ref{fig:appendix-magnitude-comparison} and~\ref{fig:appendix-fire-rate-comparison} overlay all four conditions in matched atom-rank ordering at each layer. 
Cohort SIREN distributes representation across many sparsely-firing, low-magnitude atoms at every layer.
Cohort FFMLP concentrates representation in fewer high-magnitude atoms that fire densely across the image. 
Dead atom fractions (reported per cell in the per-cell views, Figs.~\ref{fig:appendix-atom-magnitude}-\ref{fig:appendix-fire-rate}) climb to $59$-$74\%$ for FFMLP cells at layers $3$ and $4$, while cohort SIREN never exceeds $27\%$ and stays below $4\%$ from layer $1$ onward.

\begin{figure}[h]
    \centering
    \includegraphics[width=\linewidth]{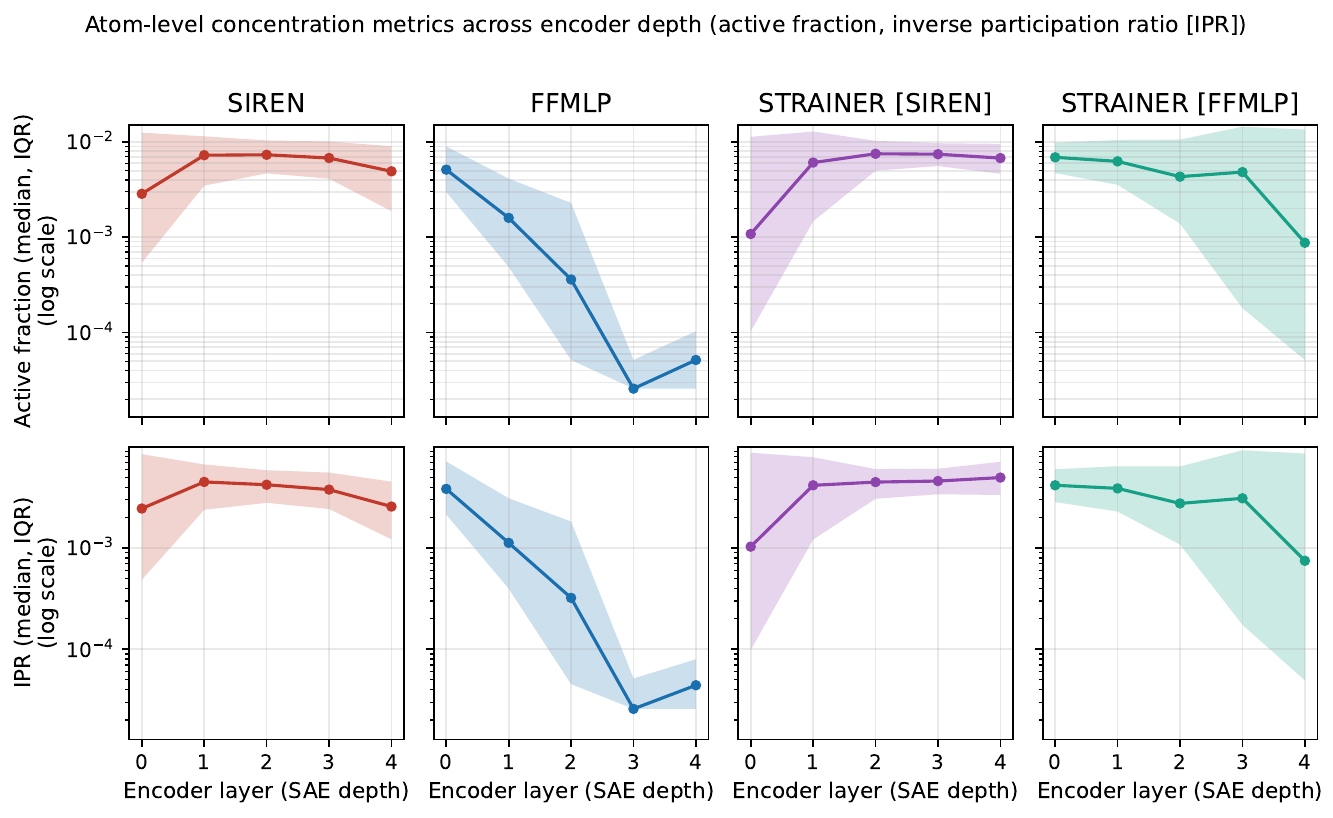}
    \caption{
        \textbf{Atom-level concentration metrics across encoder depth.}
        Top row: median active-pixel fraction ($z_a > 0.01$) over alive atoms per layer per cell. Bottom row: median inverse participation ratio (IPR) over alive atoms per layer per cell. Shaded regions show the $25$-$75\%$ range across atoms; both axes are log-scale.
        SIREN cells (red, purple) maintain stable atom-level concentration across depth in both regimes. FFMLP cells (blue, green) exhibit progressive concentration with depth: single-signal FFMLP collapses by two orders of magnitude beyond layer $1$ (median active fraction below $10^{-4}$ by layer $3$); cohort FFMLP shows the same trend more gently (factor of $\sim 7$ drop by layer $4$). The two metrics agree, confirming that the dictionary-evolution asymmetry visible in Fig.~\ref{fig:atom-gallery} corresponds to a quantitative concentration asymmetry across all cells.
    }
    \label{fig:appendix-concentration}
\end{figure}

\newpage
\clearpage

\section{Feature Galleries}
\label{sec:appendix-galleries}

\begin{figure}[h]
    \centering
    \includegraphics[width=\textwidth]{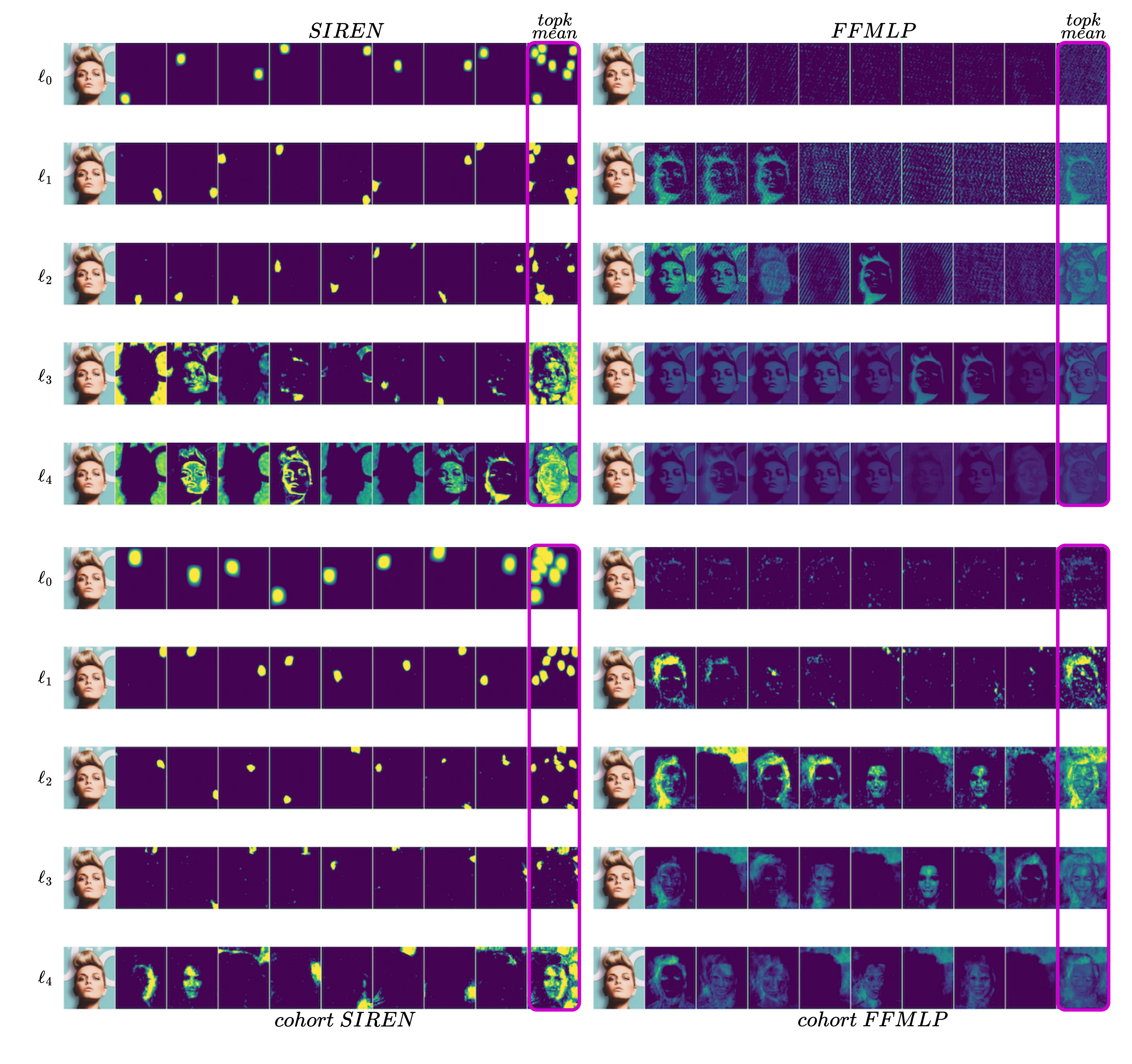}
    \caption{
        \textbf{SAE atom dictionaries across all encoder layers, CelebA cohort.}
        Top-$8$ atoms by mean magnitude at each layer $\ell_0, \ldots, \ell_4$ for each of the four conditions. The leftmost column of each block shows the input image for reference. The rightmost column (magenta border) shows the top-$k$ mean activation, the average of the top-$k$ atoms at that depth.
        Top half: single-signal SIREN (left) and single-signal FFMLP (right). Bottom half: cohort SIREN (left) and cohort FFMLP (right).
        SIREN cells display localized atoms tiling the coordinate plane at every depth, with single-signal SIREN developing image-shaped atoms only at $\ell_4$, a tendency that cohort training counteracts. FFMLP cells display image-shaped atoms that trace memorized cohort contours from $\ell_2$ onward, with the contours intensifying with depth in cohort FFMLP. The dichotomy in Fig.~\ref{fig:atom-gallery} (main text) holds across all encoder depths.
    }
    \label{fig:appendix-gallery-all-layers}
\end{figure}

\begin{figure}[h]
    \centering
    \includegraphics[width=\textwidth]{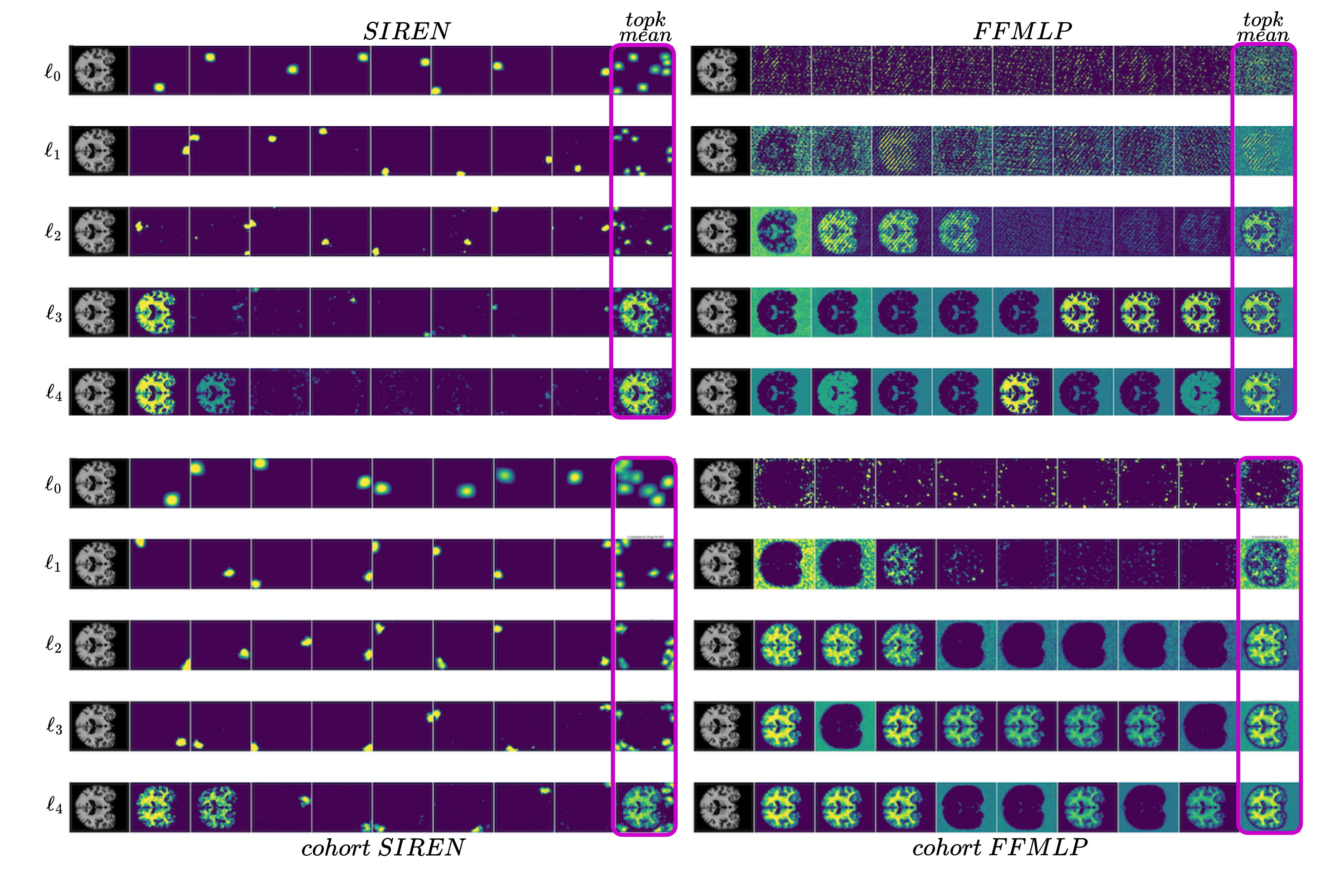}
    \caption{
        \textbf{SAE atom dictionaries across all encoder layers, OASIS cohort.}
        Top-$8$ atoms by mean magnitude at each layer $\ell_0, \ldots, \ell_4$ for each of the four conditions. The leftmost column of each block shows the input image for reference. The rightmost column (magenta border) shows the top-$k$ mean activation, the average of the top-$k$ atoms at that depth.
        Top half: single-signal SIREN (left) and single-signal FFMLP (right). Bottom half: cohort SIREN (left) and cohort FFMLP (right).
        SIREN cells display localized atoms tiling the coordinate plane across all depths, with single-signal SIREN developing image-shaped atoms only at $\ell_3$ and $\ell_4$, a tendency that cohort training counteracts. FFMLP cells display image-shaped atoms tracing the brain anatomy from $\ell_2$ onward, with the cohort variant showing sharper, more saturated brain contours than the single-signal variant. The same architectural dichotomy observed on the CelebA cohort (Fig.~\ref{fig:appendix-gallery-all-layers}) holds on OASIS, confirming that the split is driven by architecture rather than cohort domain.
    }
    \label{fig:appendix-gallery-all-layers-oasis}
\end{figure}

\newpage
\clearpage

\section{Per-layer atom ablations}
\label{sec:appendix-ablations}

\begin{figure}[h]
    \centering
    \includegraphics[width=\textwidth]{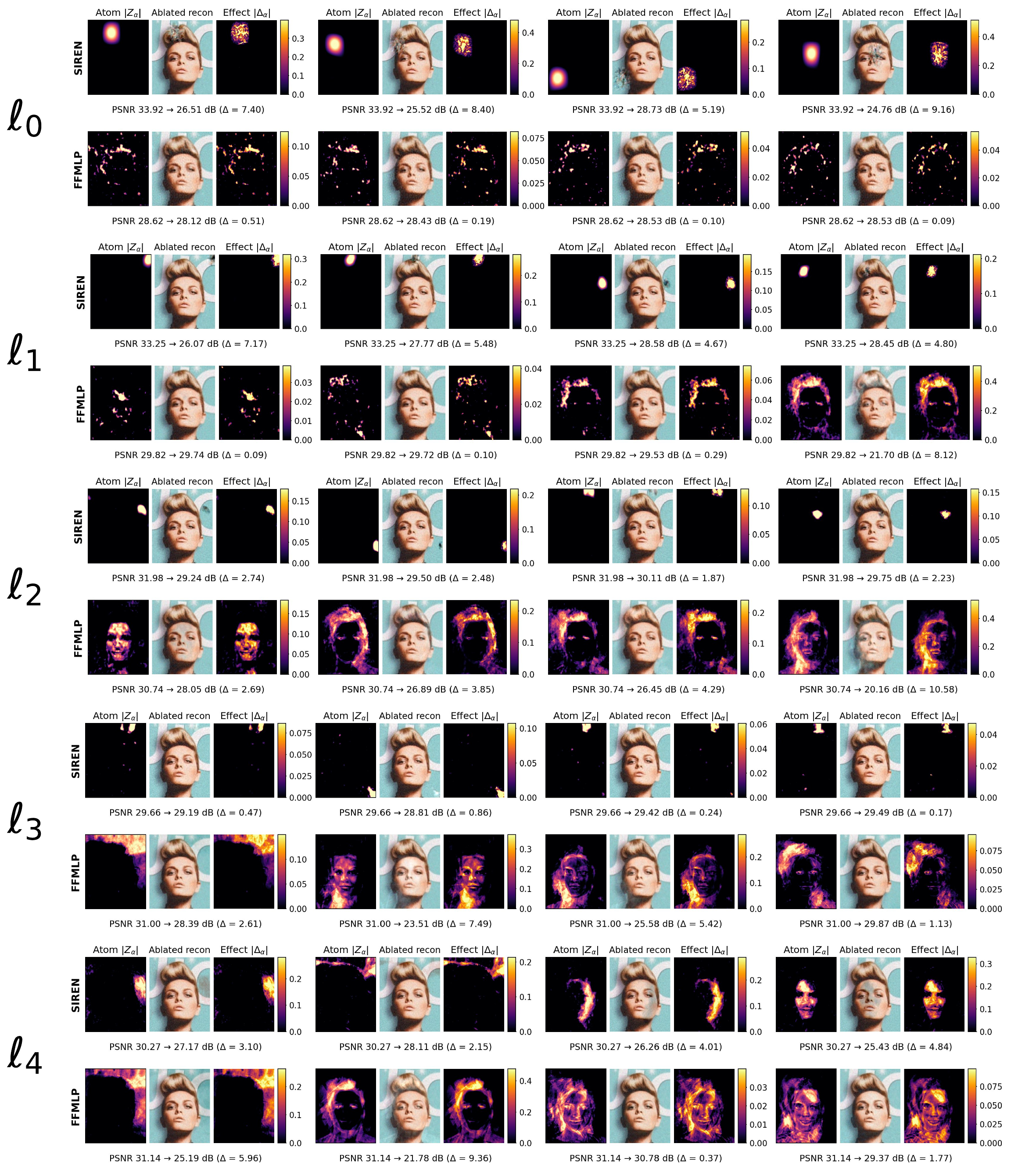}
    \caption{
        \textbf{Single-atom ablations across all encoder layers for cohort SIREN and cohort FFMLP.}
        Top-$4$ atoms by mean magnitude at each layer $\ell_0, \ldots, \ell_4$, ablated on a CelebA test image. For each atom: \emph{Atom $|Z_a|$} shows where the atom fires, \emph{Ablated recon} the network output after zeroing the atom, and \emph{Effect $|\Delta_a|$} the per-pixel change in reconstruction. PSNR before $\to$ after ablation reported below each example. SIREN effects stay confined to the atom's firing region across all depths, with PSNR drops of $0.2$-$9.2$ dB localized to that region. FFMLP effects spread across the entire image, with single-atom PSNR drops up to $10.58$ dB at $\ell_2$ and $9.36$ dB at $\ell_4$. The asymmetry observed at $\ell_2$ in Fig.~\ref{fig:atom-ablation} (main text) persists at all encoder depths.
    }
    \label{fig:ablation-all-layers}
\end{figure}

\newpage
\clearpage

\section{Comparison to STRAINER}
\label{sec:appendix-strainer-comparison}

\begin{table}[h]
\centering
\small
\setlength{\tabcolsep}{4pt}
\caption{Full experimental setup compared to STRAINER~\citep{vyas2025learning}. Our protocol matches STRAINER's published recipe except where noted. Backbone-specific hyperparameters ($\omega_0$ for SIREN, $\sigma$ for FFMLP) are tuned per regime on a held-out validation set of $50$ images for CelebA-HQ and OASIS, and reused for Kodak. All experiments were performed on a workstation with an AMD Ryzen\textsuperscript{TM} Threadripper 3960X CPU and NVIDIA Quadro RTX 8000 GPU.}
\label{tab:strainer-vs-ours}
\begin{tabular}{l l l}
\toprule
 & \textbf{STRAINER} & \textbf{Ours} \\
\midrule
\multicolumn{3}{l}{\textit{Architecture}} \\
Backbone                  & SIREN                       & SIREN, FFMLP \\
Shared encoder layers $L$ & 5                           & 5 \\
Per-signal head layers    & 1                           & 1 \\
Hidden width $d$          & 256                         & 256 \\
SIREN initialization      & standard~\citep{sitzmann2020implicit} & standard~\citep{sitzmann2020implicit} \\
FFMLP features            & ---                         & $F = 256$, $\mathcal{N}(0, \sigma^2)$, fixed \\
FFMLP nonlinearity        & ---                         & ReLU \\
\midrule
\multicolumn{3}{l}{\textit{Single-signal training (per image)}} \\
Iterations                & ---                         & 500 \\
Optimizer                 & ---                         & Adam~\citep{kingma2014adam} \\
SIREN $\omega_0$ / lr     & ---                         & $50$ / $2 \times 10^{-4}$ \\
FFMLP $\sigma$ / lr       & ---                         & $20$ / $5 \times 10^{-4}$ \\
\midrule
\multicolumn{3}{l}{\textit{Cohort training}} \\
Cohort size $M$           & 10                          & 10 \\
Pretraining iterations    & 5000                        & 5000 \\
Pretraining stop crit.    & cohort PSNR $\approx$ 30 dB & cohort PSNR $\approx$ 30 dB \\
Optimizer                 & Adam                        & Adam \\
Learning rate (both)      & $10^{-4}$                   & $10^{-4}$ \\
SIREN $\omega_0$          & $30$               & $30$ \\
FFMLP $\sigma$            & ---                         & $10$ \\
\midrule
\multicolumn{3}{l}{\textit{Test-time fitting (cohort-trained)}} \\
Iterations                & 2000                        & 500 \\
Optimizer                 & Adam                        & Adam \\
Learning rate             & $10^{-4}$                   & $10^{-4}$ \\
Coordinate range          & $[-1, 1]$                   & $[-1, 1]$ \\
Pixel range               & $[0, 1]$                    & $[0, 1]$ \\
Reinitialization seeds    & not reported                & 3 \\
\midrule
\multicolumn{3}{l}{\textit{Datasets and resolutions}} \\
CelebA-HQ~\citep{karras2017progressive} & $178 \times 178$ & $178 \times 218$ \\
OASIS-MRI~\citep{Marcus2007OpenAS,hoopes2022learning}      & not reported      & $160 \times 192$ \\
Kodak~\citep{kodak1993kodak}            & $768 \times 512$  & $768 \times 512$ \\
\midrule
\multicolumn{3}{l}{\textit{Test split sizes}} \\
CelebA-HQ            & $550$                       & $100$ \\
OASIS-MRI            & $144$                       & $100$ \\
Kodak                & $3$-$4$ named images        & $14$ \\
AFHQ                 & $368$                       & not used \\
\midrule
\multicolumn{3}{l}{\textit{Cohort encoder per target}} \\
Cohort resolution         & matched to target           & native source resolution \\
\midrule
\multicolumn{3}{l}{\textit{Test-time protocol}} \\
Encoder weights           & init from pretrained        & frozen up to $\tau$, init beyond \\
Decoder weights           & random init                 & random init \\
Optimization              & all weights updated         & only unfrozen weights updated \\
\bottomrule
\end{tabular}
\end{table}

\begin{table*}[t]
\centering
\scriptsize
\setlength{\tabcolsep}{2.5pt}
\caption{Effect of freeze depth on transfer for \textbf{CelebA}-pretrained STRAINER, with STRAINER [\textbf{SIREN}] (left) and STRAINER [\textbf{FFMLP}] (right) shown side-by-side per target distribution. \texttt{sr(w)} and \texttt{sr(h)} are pre-fit weight stable rank and pre-fit activation (hidden) stable rank at the freeze-boundary layer $\tau$ (constant per checkpoint). The layer with the maximum \texttt{sr(w)} is highlighted in \textbf{bold} (both the $\tau$ index and the \texttt{sr(w)} value). Metrics are mean $\pm$ std across images. Best PSNR row per (cell, arch) in \textbf{bold}; layer 4 collapses for every cell. The row marked \emph{*} (separated by a horizontal line) reports the published STRAINER recipe (pretrained encoder, retrained on each test image). The right block aligns row-by-row with the $\tau$ column of the left block.}
\label{tab:strainer-celeba}

\noindent\textit{CelebA $\to$ CelebA (in-dist)}\par
\vspace{2pt}
\begin{minipage}[t]{0.55\textwidth}
\centering STRAINER [\textbf{SIREN}]\\[2pt]
\begin{tabular}{cccccc}
\toprule
$\tau$ & sr(w) & sr(h) & PSNR (dB) $\uparrow$ & SSIM $\uparrow$ & LPIPS $\downarrow$ \\
\midrule
* & -- & -- & 43.15{\scriptsize $\pm$ 2.26} & 0.9879{\scriptsize $\pm$ 0.0027} & 0.0035{\scriptsize $\pm$ 0.0017} \\
\cmidrule(lr){1-6}
0 & 1.97 & 14.91 & 41.09{\scriptsize $\pm$ 2.00} & 0.9814{\scriptsize $\pm$ 0.0066} & 0.0213{\scriptsize $\pm$ 0.0161} \\
\textbf{1} & \textbf{32.45} & 32.26 & \textbf{42.36{\scriptsize $\pm$ 2.47}} & \textbf{0.9858{\scriptsize $\pm$ 0.0052}} & \textbf{0.0124{\scriptsize $\pm$ 0.0097}} \\
2 & 28.04 & 48.55 & 41.65{\scriptsize $\pm$ 2.61} & 0.9845{\scriptsize $\pm$ 0.0060} & 0.0120{\scriptsize $\pm$ 0.0094} \\
3 & 29.16 & 39.43 & 38.66{\scriptsize $\pm$ 3.28} & 0.9695{\scriptsize $\pm$ 0.0123} & 0.0203{\scriptsize $\pm$ 0.0096} \\
4 & 21.72 & 16.19 & 17.57{\scriptsize $\pm$ 1.39} & 0.3295{\scriptsize $\pm$ 0.0531} & 0.5498{\scriptsize $\pm$ 0.0552} \\
\bottomrule
\end{tabular}
\end{minipage}
\hfill
\begin{minipage}[t]{0.43\textwidth}
\centering STRAINER [\textbf{FFMLP}]\\[2pt]
\begin{tabular}{ccccc}
\toprule
sr(w) & sr(h) & PSNR (dB) $\uparrow$ & SSIM $\uparrow$ & LPIPS $\downarrow$ \\
\midrule
-- & -- & 33.20{\scriptsize $\pm$ 2.42} & 0.9016{\scriptsize $\pm$ 0.0237} & 0.1093{\scriptsize $\pm$ 0.0317} \\
\cmidrule(lr){1-5}
\textbf{79.61} & 63.08 & \textbf{34.01{\scriptsize $\pm$ 2.40}} & \textbf{0.9148{\scriptsize $\pm$ 0.0218}} & \textbf{0.0847{\scriptsize $\pm$ 0.0308}} \\
30.59 & 11.17 & 29.79{\scriptsize $\pm$ 2.16} & 0.8098{\scriptsize $\pm$ 0.0408} & 0.2243{\scriptsize $\pm$ 0.0583} \\
16.66 & 3.54 & 25.50{\scriptsize $\pm$ 1.69} & 0.6472{\scriptsize $\pm$ 0.0521} & 0.3850{\scriptsize $\pm$ 0.0622} \\
23.11 & 2.45 & 20.71{\scriptsize $\pm$ 1.52} & 0.4799{\scriptsize $\pm$ 0.0629} & 0.4681{\scriptsize $\pm$ 0.0573} \\
26.76 & 2.28 & 15.92{\scriptsize $\pm$ 1.37} & 0.3913{\scriptsize $\pm$ 0.0616} & 0.5077{\scriptsize $\pm$ 0.0594} \\
\bottomrule
\end{tabular}
\end{minipage}
\vspace{8pt}

\noindent\textit{CelebA $\to$ Kodak (OOD)}\par
\vspace{2pt}
\begin{minipage}[t]{0.55\textwidth}
\centering STRAINER [\textbf{SIREN}]\\[2pt]
\begin{tabular}{cccccc}
\toprule
$\tau$ & sr(w) & sr(h) & PSNR (dB) $\uparrow$ & SSIM $\uparrow$ & LPIPS $\downarrow$ \\
\midrule
* & -- & -- & 30.93{\scriptsize $\pm$ 2.65} & 0.8403{\scriptsize $\pm$ 0.0466} & 0.2295{\scriptsize $\pm$ 0.0561} \\
\cmidrule(lr){1-6}
0 & 1.97 & 14.91 & 37.89{\scriptsize $\pm$ 2.28} & 0.9641{\scriptsize $\pm$ 0.0117} & 0.0248{\scriptsize $\pm$ 0.0149} \\
\textbf{1} & \textbf{32.45} & 32.26 & \textbf{39.77{\scriptsize $\pm$ 1.99}} & \textbf{0.9754{\scriptsize $\pm$ 0.0065}} & \textbf{0.0111{\scriptsize $\pm$ 0.0067}} \\
2 & 28.04 & 48.55 & 37.68{\scriptsize $\pm$ 2.16} & 0.9633{\scriptsize $\pm$ 0.0076} & 0.0166{\scriptsize $\pm$ 0.0063} \\
3 & 29.16 & 39.43 & 32.14{\scriptsize $\pm$ 2.49} & 0.9004{\scriptsize $\pm$ 0.0176} & 0.1054{\scriptsize $\pm$ 0.0228} \\
4 & 21.72 & 16.19 & 17.87{\scriptsize $\pm$ 1.60} & 0.3302{\scriptsize $\pm$ 0.0950} & 0.8385{\scriptsize $\pm$ 0.0616} \\
\bottomrule
\end{tabular}
\end{minipage}
\hfill
\begin{minipage}[t]{0.43\textwidth}
\centering STRAINER [\textbf{FFMLP}]\\[2pt]
\begin{tabular}{ccccc}
\toprule
sr(w) & sr(h) & PSNR (dB) $\uparrow$ & SSIM $\uparrow$ & LPIPS $\downarrow$ \\
\midrule
-- & -- & 28.43{\scriptsize $\pm$ 2.54} & 0.7841{\scriptsize $\pm$ 0.0623} & 0.3013{\scriptsize $\pm$ 0.0524} \\
\cmidrule(lr){1-5}
\textbf{79.61} & 63.08 & \textbf{29.89{\scriptsize $\pm$ 3.78}} & \textbf{0.8215{\scriptsize $\pm$ 0.1197}} & \textbf{0.2313{\scriptsize $\pm$ 0.1366}} \\
30.59 & 11.17 & 26.87{\scriptsize $\pm$ 3.32} & 0.7110{\scriptsize $\pm$ 0.1168} & 0.4053{\scriptsize $\pm$ 0.1384} \\
16.66 & 3.54 & 23.85{\scriptsize $\pm$ 2.72} & 0.5647{\scriptsize $\pm$ 0.1091} & 0.6104{\scriptsize $\pm$ 0.1066} \\
23.11 & 2.45 & 20.90{\scriptsize $\pm$ 2.19} & 0.4437{\scriptsize $\pm$ 0.1123} & 0.7365{\scriptsize $\pm$ 0.0822} \\
26.76 & 2.28 & 17.40{\scriptsize $\pm$ 1.69} & 0.4042{\scriptsize $\pm$ 0.1281} & 0.8240{\scriptsize $\pm$ 0.0892} \\
\bottomrule
\end{tabular}
\end{minipage}
\vspace{8pt}

\noindent\textit{CelebA $\to$ OASIS (OOD)}\par
\vspace{2pt}
\begin{minipage}[t]{0.55\textwidth}
\centering STRAINER [\textbf{SIREN}]\\[2pt]
\begin{tabular}{cccccc}
\toprule
$\tau$ & sr(w) & sr(h) & PSNR (dB) $\uparrow$ & SSIM $\uparrow$ & LPIPS $\downarrow$ \\
\midrule
* & -- & -- & 49.28{\scriptsize $\pm$ 1.30} & 0.9914{\scriptsize $\pm$ 0.0022} & 0.0007{\scriptsize $\pm$ 0.0002} \\
\cmidrule(lr){1-6}
0 & 1.97 & 14.91 & 48.62{\scriptsize $\pm$ 1.50} & 0.9938{\scriptsize $\pm$ 0.0078} & 0.0031{\scriptsize $\pm$ 0.0020} \\
\textbf{1} & \textbf{32.45} & 32.26 & \textbf{49.52{\scriptsize $\pm$ 1.42}} & \textbf{0.9971{\scriptsize $\pm$ 0.0005}} & \textbf{0.0019{\scriptsize $\pm$ 0.0007}} \\
2 & 28.04 & 48.55 & 48.76{\scriptsize $\pm$ 1.71} & 0.9960{\scriptsize $\pm$ 0.0010} & 0.0017{\scriptsize $\pm$ 0.0007} \\
3 & 29.16 & 39.43 & 45.24{\scriptsize $\pm$ 1.95} & 0.9900{\scriptsize $\pm$ 0.0034} & 0.0038{\scriptsize $\pm$ 0.0012} \\
4 & 21.72 & 16.19 & 18.36{\scriptsize $\pm$ 1.68} & 0.2516{\scriptsize $\pm$ 0.0357} & 0.4748{\scriptsize $\pm$ 0.0189} \\
\bottomrule
\end{tabular}
\end{minipage}
\hfill
\begin{minipage}[t]{0.43\textwidth}
\centering STRAINER [\textbf{FFMLP}]\\[2pt]
\begin{tabular}{ccccc}
\toprule
sr(w) & sr(h) & PSNR (dB) $\uparrow$ & SSIM $\uparrow$ & LPIPS $\downarrow$ \\
\midrule
-- & -- & 39.61{\scriptsize $\pm$ 1.57} & 0.9468{\scriptsize $\pm$ 0.0111} & 0.0222{\scriptsize $\pm$ 0.0056} \\
\cmidrule(lr){1-5}
\textbf{79.61} & 63.08 & \textbf{42.93{\scriptsize $\pm$ 2.21}} & \textbf{0.9894{\scriptsize $\pm$ 0.0023}} & \textbf{0.0050{\scriptsize $\pm$ 0.0018}} \\
30.59 & 11.17 & 35.76{\scriptsize $\pm$ 1.94} & 0.9239{\scriptsize $\pm$ 0.0180} & 0.0545{\scriptsize $\pm$ 0.0119} \\
16.66 & 3.54 & 28.85{\scriptsize $\pm$ 1.75} & 0.7030{\scriptsize $\pm$ 0.0489} & 0.2534{\scriptsize $\pm$ 0.0199} \\
23.11 & 2.45 & 22.35{\scriptsize $\pm$ 1.83} & 0.4681{\scriptsize $\pm$ 0.0496} & 0.3913{\scriptsize $\pm$ 0.0214} \\
26.76 & 2.28 & 17.04{\scriptsize $\pm$ 1.88} & 0.2765{\scriptsize $\pm$ 0.0442} & 0.4626{\scriptsize $\pm$ 0.0205} \\
\bottomrule
\end{tabular}
\end{minipage}
\vspace{8pt}

\end{table*}

\begin{table*}[t]
\centering
\scriptsize
\setlength{\tabcolsep}{2.5pt}
\caption{Effect of freeze depth on transfer for \textbf{Kodak}-pretrained STRAINER, with STRAINER [\textbf{SIREN}] (left) and STRAINER [\textbf{FFMLP}] (right) shown side-by-side per target distribution. \texttt{sr(w)} and \texttt{sr(h)} are pre-fit weight stable rank and pre-fit activation (hidden) stable rank at the freeze-boundary layer $\tau$ (constant per checkpoint). The layer with the maximum \texttt{sr(w)} is highlighted in \textbf{bold} (both the $\tau$ index and the \texttt{sr(w)} value). Metrics are mean $\pm$ std across images. Best PSNR row per (cell, arch) in \textbf{bold}; layer 4 collapses for every cell. The row marked \emph{*} (separated by a horizontal line) reports the published STRAINER recipe (pretrained encoder, retrained on each test image). The right block aligns row-by-row with the $\tau$ column of the left block.}
\label{tab:strainer-kodak}

\noindent\textit{Kodak $\to$ CelebA (OOD)}\par
\vspace{2pt}
\begin{minipage}[t]{0.55\textwidth}
\centering STRAINER [\textbf{SIREN}]\\[2pt]
\begin{tabular}{cccccc}
\toprule
$\tau$ & sr(w) & sr(h) & PSNR (dB) $\uparrow$ & SSIM $\uparrow$ & LPIPS $\downarrow$ \\
\midrule
* & -- & -- & 43.41{\scriptsize $\pm$ 2.08} & 0.9876{\scriptsize $\pm$ 0.0030} & 0.0028{\scriptsize $\pm$ 0.0013} \\
\cmidrule(lr){1-6}
0 & 1.98 & 15.14 & 40.97{\scriptsize $\pm$ 2.35} & 0.9784{\scriptsize $\pm$ 0.0130} & 0.0223{\scriptsize $\pm$ 0.0147} \\
\textbf{1} & \textbf{33.62} & 33.94 & \textbf{42.67{\scriptsize $\pm$ 2.21}} & \textbf{0.9868{\scriptsize $\pm$ 0.0053}} & \textbf{0.0116{\scriptsize $\pm$ 0.0104}} \\
2 & 29.80 & 50.32 & 41.98{\scriptsize $\pm$ 2.44} & 0.9855{\scriptsize $\pm$ 0.0055} & 0.0100{\scriptsize $\pm$ 0.0088} \\
3 & 24.55 & 53.73 & 38.91{\scriptsize $\pm$ 3.23} & 0.9698{\scriptsize $\pm$ 0.0120} & 0.0164{\scriptsize $\pm$ 0.0082} \\
4 & 20.78 & 31.25 & 15.40{\scriptsize $\pm$ 1.06} & 0.2303{\scriptsize $\pm$ 0.0350} & 0.6826{\scriptsize $\pm$ 0.0480} \\
\bottomrule
\end{tabular}
\end{minipage}
\hfill
\begin{minipage}[t]{0.43\textwidth}
\centering STRAINER [\textbf{FFMLP}]\\[2pt]
\begin{tabular}{ccccc}
\toprule
sr(w) & sr(h) & PSNR (dB) $\uparrow$ & SSIM $\uparrow$ & LPIPS $\downarrow$ \\
\midrule
-- & -- & 33.61{\scriptsize $\pm$ 2.23} & 0.9088{\scriptsize $\pm$ 0.0181} & 0.0936{\scriptsize $\pm$ 0.0234} \\
\cmidrule(lr){1-5}
\textbf{81.07} & 65.84 & \textbf{33.83{\scriptsize $\pm$ 2.35}} & \textbf{0.9078{\scriptsize $\pm$ 0.0231}} & \textbf{0.0877{\scriptsize $\pm$ 0.0261}} \\
40.84 & 23.73 & 29.27{\scriptsize $\pm$ 1.86} & 0.7831{\scriptsize $\pm$ 0.0381} & 0.2513{\scriptsize $\pm$ 0.0513} \\
20.87 & 6.64 & 24.18{\scriptsize $\pm$ 1.42} & 0.5649{\scriptsize $\pm$ 0.0396} & 0.4952{\scriptsize $\pm$ 0.0506} \\
18.90 & 3.52 & 17.73{\scriptsize $\pm$ 1.28} & 0.3380{\scriptsize $\pm$ 0.0354} & 0.6701{\scriptsize $\pm$ 0.0514} \\
24.09 & 3.29 & 14.06{\scriptsize $\pm$ 1.13} & 0.3169{\scriptsize $\pm$ 0.0602} & 0.7453{\scriptsize $\pm$ 0.0475} \\
\bottomrule
\end{tabular}
\end{minipage}
\vspace{8pt}

\noindent\textit{Kodak $\to$ Kodak (in-dist)}\par
\vspace{2pt}
\begin{minipage}[t]{0.55\textwidth}
\centering STRAINER [\textbf{SIREN}]\\[2pt]
\begin{tabular}{cccccc}
\toprule
$\tau$ & sr(w) & sr(h) & PSNR (dB) $\uparrow$ & SSIM $\uparrow$ & LPIPS $\downarrow$ \\
\midrule
* & -- & -- & 30.02{\scriptsize $\pm$ 2.63} & 0.8485{\scriptsize $\pm$ 0.0357} & 0.2174{\scriptsize $\pm$ 0.0541} \\
\cmidrule(lr){1-6}
0 & 1.98 & 15.14 & 38.05{\scriptsize $\pm$ 2.36} & 0.9681{\scriptsize $\pm$ 0.0083} & 0.0205{\scriptsize $\pm$ 0.0083} \\
\textbf{1} & \textbf{33.62} & 33.94 & \textbf{39.81{\scriptsize $\pm$ 2.11}} & \textbf{0.9767{\scriptsize $\pm$ 0.0066}} & \textbf{0.0089{\scriptsize $\pm$ 0.0036}} \\
2 & 29.80 & 50.32 & 38.01{\scriptsize $\pm$ 2.37} & 0.9659{\scriptsize $\pm$ 0.0078} & 0.0136{\scriptsize $\pm$ 0.0046} \\
3 & 24.55 & 53.73 & 33.58{\scriptsize $\pm$ 2.45} & 0.9190{\scriptsize $\pm$ 0.0149} & 0.0674{\scriptsize $\pm$ 0.0195} \\
4 & 20.78 & 31.25 & 29.95{\scriptsize $\pm$ 1.64} & 0.8332{\scriptsize $\pm$ 0.0284} & 0.2074{\scriptsize $\pm$ 0.0734} \\
\bottomrule
\end{tabular}
\end{minipage}
\hfill
\begin{minipage}[t]{0.43\textwidth}
\centering STRAINER [\textbf{FFMLP}]\\[2pt]
\begin{tabular}{ccccc}
\toprule
sr(w) & sr(h) & PSNR (dB) $\uparrow$ & SSIM $\uparrow$ & LPIPS $\downarrow$ \\
\midrule
-- & -- & 28.76{\scriptsize $\pm$ 2.62} & 0.8214{\scriptsize $\pm$ 0.0501} & 0.2620{\scriptsize $\pm$ 0.0586} \\
\cmidrule(lr){1-5}
\textbf{81.07} & 65.84 & \textbf{31.06{\scriptsize $\pm$ 2.78}} & \textbf{0.8663{\scriptsize $\pm$ 0.0310}} & \textbf{0.1620{\scriptsize $\pm$ 0.0351}} \\
40.84 & 23.73 & 29.24{\scriptsize $\pm$ 2.29} & 0.8149{\scriptsize $\pm$ 0.0369} & 0.2528{\scriptsize $\pm$ 0.0632} \\
20.87 & 6.64 & 27.87{\scriptsize $\pm$ 1.70} & 0.7638{\scriptsize $\pm$ 0.0653} & 0.3416{\scriptsize $\pm$ 0.1074} \\
18.90 & 3.52 & 26.98{\scriptsize $\pm$ 2.09} & 0.7308{\scriptsize $\pm$ 0.1013} & 0.4029{\scriptsize $\pm$ 0.1381} \\
24.09 & 3.29 & 25.30{\scriptsize $\pm$ 2.60} & 0.7037{\scriptsize $\pm$ 0.1105} & 0.4838{\scriptsize $\pm$ 0.1615} \\
\bottomrule
\end{tabular}
\end{minipage}
\vspace{8pt}

\noindent\textit{Kodak $\to$ OASIS (OOD)}\par
\vspace{2pt}
\begin{minipage}[t]{0.55\textwidth}
\centering STRAINER [\textbf{SIREN}]\\[2pt]
\begin{tabular}{cccccc}
\toprule
$\tau$ & sr(w) & sr(h) & PSNR (dB) $\uparrow$ & SSIM $\uparrow$ & LPIPS $\downarrow$ \\
\midrule
* & -- & -- & 49.78{\scriptsize $\pm$ 1.00} & 0.9941{\scriptsize $\pm$ 0.0017} & 0.0008{\scriptsize $\pm$ 0.0003} \\
\cmidrule(lr){1-6}
0 & 1.98 & 15.14 & 49.09{\scriptsize $\pm$ 1.35} & 0.9962{\scriptsize $\pm$ 0.0020} & 0.0024{\scriptsize $\pm$ 0.0010} \\
\textbf{1} & \textbf{33.62} & 33.94 & \textbf{49.83{\scriptsize $\pm$ 1.42}} & \textbf{0.9956{\scriptsize $\pm$ 0.0062}} & \textbf{0.0017{\scriptsize $\pm$ 0.0006}} \\
2 & 29.80 & 50.32 & 49.02{\scriptsize $\pm$ 1.61} & 0.9959{\scriptsize $\pm$ 0.0014} & 0.0013{\scriptsize $\pm$ 0.0006} \\
3 & 24.55 & 53.73 & 47.50{\scriptsize $\pm$ 1.95} & 0.9929{\scriptsize $\pm$ 0.0024} & 0.0012{\scriptsize $\pm$ 0.0004} \\
4 & 20.78 & 31.25 & 17.02{\scriptsize $\pm$ 1.70} & 0.1685{\scriptsize $\pm$ 0.0277} & 0.5759{\scriptsize $\pm$ 0.0176} \\
\bottomrule
\end{tabular}
\end{minipage}
\hfill
\begin{minipage}[t]{0.43\textwidth}
\centering STRAINER [\textbf{FFMLP}]\\[2pt]
\begin{tabular}{ccccc}
\toprule
sr(w) & sr(h) & PSNR (dB) $\uparrow$ & SSIM $\uparrow$ & LPIPS $\downarrow$ \\
\midrule
-- & -- & 40.82{\scriptsize $\pm$ 1.49} & 0.9648{\scriptsize $\pm$ 0.0089} & 0.0143{\scriptsize $\pm$ 0.0030} \\
\cmidrule(lr){1-5}
\textbf{81.07} & 65.84 & \textbf{43.84{\scriptsize $\pm$ 1.87}} & \textbf{0.9894{\scriptsize $\pm$ 0.0027}} & \textbf{0.0034{\scriptsize $\pm$ 0.0011}} \\
40.84 & 23.73 & 36.34{\scriptsize $\pm$ 1.60} & 0.9212{\scriptsize $\pm$ 0.0166} & 0.0530{\scriptsize $\pm$ 0.0123} \\
20.87 & 6.64 & 28.54{\scriptsize $\pm$ 1.56} & 0.6058{\scriptsize $\pm$ 0.0376} & 0.3109{\scriptsize $\pm$ 0.0176} \\
18.90 & 3.52 & 20.59{\scriptsize $\pm$ 1.82} & 0.3035{\scriptsize $\pm$ 0.0404} & 0.5114{\scriptsize $\pm$ 0.0189} \\
24.09 & 3.29 & 15.67{\scriptsize $\pm$ 1.80} & 0.1407{\scriptsize $\pm$ 0.0304} & 0.5650{\scriptsize $\pm$ 0.0282} \\
\bottomrule
\end{tabular}
\end{minipage}
\vspace{8pt}

\end{table*}

\begin{table*}[t]
\centering
\scriptsize
\setlength{\tabcolsep}{2.5pt}
\caption{Effect of freeze depth on transfer for \textbf{OASIS}-pretrained STRAINER, with STRAINER [\textbf{SIREN}] (left) and STRAINER [\textbf{FFMLP}] (right) shown side-by-side per target distribution. \texttt{sr(w)} and \texttt{sr(h)} are pre-fit weight stable rank and pre-fit activation (hidden) stable rank at the freeze-boundary layer $\tau$ (constant per checkpoint). The layer with the maximum \texttt{sr(w)} is highlighted in \textbf{bold} (both the $\tau$ index and the \texttt{sr(w)} value). Metrics are mean $\pm$ std across images. Best PSNR row per (cell, arch) in \textbf{bold}; layer 4 collapses for every cell. The row marked \emph{*} (separated by a horizontal line) reports the published STRAINER recipe (pretrained encoder, retrained on each test image). The right block aligns row-by-row with the $\tau$ column of the left block.}
\label{tab:strainer-oasis}

\noindent\textit{OASIS $\to$ CelebA (OOD)}\par
\vspace{2pt}
\begin{minipage}[t]{0.55\textwidth}
\centering STRAINER [\textbf{SIREN}]\\[2pt]
\begin{tabular}{cccccc}
\toprule
$\tau$ & sr(w) & sr(h) & PSNR (dB) $\uparrow$ & SSIM $\uparrow$ & LPIPS $\downarrow$ \\
\midrule
* & -- & -- & 40.12{\scriptsize $\pm$ 3.34} & 0.9812{\scriptsize $\pm$ 0.0091} & 0.0208{\scriptsize $\pm$ 0.0135} \\
\cmidrule(lr){1-6}
0 & 1.97 & 14.54 & 41.04{\scriptsize $\pm$ 2.23} & 0.9813{\scriptsize $\pm$ 0.0063} & 0.0205{\scriptsize $\pm$ 0.0142} \\
\textbf{1} & \textbf{31.89} & 17.54 & \textbf{41.87{\scriptsize $\pm$ 2.15}} & \textbf{0.9842{\scriptsize $\pm$ 0.0060}} & \textbf{0.0145{\scriptsize $\pm$ 0.0123}} \\
2 & 28.22 & 12.89 & 40.25{\scriptsize $\pm$ 2.70} & 0.9798{\scriptsize $\pm$ 0.0082} & 0.0173{\scriptsize $\pm$ 0.0138} \\
3 & 25.30 & 8.66 & 36.20{\scriptsize $\pm$ 3.23} & 0.9554{\scriptsize $\pm$ 0.0183} & 0.0435{\scriptsize $\pm$ 0.0190} \\
4 & 25.13 & 4.89 & 15.27{\scriptsize $\pm$ 0.96} & 0.2665{\scriptsize $\pm$ 0.0319} & 0.6282{\scriptsize $\pm$ 0.0496} \\
\bottomrule
\end{tabular}
\end{minipage}
\hfill
\begin{minipage}[t]{0.43\textwidth}
\centering STRAINER [\textbf{FFMLP}]\\[2pt]
\begin{tabular}{ccccc}
\toprule
sr(w) & sr(h) & PSNR (dB) $\uparrow$ & SSIM $\uparrow$ & LPIPS $\downarrow$ \\
\midrule
-- & -- & 34.71{\scriptsize $\pm$ 3.15} & 0.9315{\scriptsize $\pm$ 0.0233} & 0.1071{\scriptsize $\pm$ 0.0407} \\
\cmidrule(lr){1-5}
\textbf{66.77} & 33.32 & \textbf{33.21{\scriptsize $\pm$ 2.30}} & \textbf{0.8933{\scriptsize $\pm$ 0.0260}} & \textbf{0.1067{\scriptsize $\pm$ 0.0309}} \\
26.94 & 4.93 & 28.95{\scriptsize $\pm$ 1.77} & 0.7611{\scriptsize $\pm$ 0.0363} & 0.2694{\scriptsize $\pm$ 0.0443} \\
21.92 & 1.52 & 22.40{\scriptsize $\pm$ 1.86} & 0.5085{\scriptsize $\pm$ 0.0520} & 0.5220{\scriptsize $\pm$ 0.0533} \\
20.81 & 1.13 & 15.00{\scriptsize $\pm$ 1.84} & 0.3235{\scriptsize $\pm$ 0.0543} & 0.6631{\scriptsize $\pm$ 0.0651} \\
22.94 & 1.10 & 12.79{\scriptsize $\pm$ 1.37} & 0.3243{\scriptsize $\pm$ 0.0647} & 0.6365{\scriptsize $\pm$ 0.0587} \\
\bottomrule
\end{tabular}
\end{minipage}
\vspace{8pt}

\noindent\textit{OASIS $\to$ Kodak (OOD)}\par
\vspace{2pt}
\begin{minipage}[t]{0.55\textwidth}
\centering STRAINER [\textbf{SIREN}]\\[2pt]
\begin{tabular}{cccccc}
\toprule
$\tau$ & sr(w) & sr(h) & PSNR (dB) $\uparrow$ & SSIM $\uparrow$ & LPIPS $\downarrow$ \\
\midrule
* & -- & -- & 30.56{\scriptsize $\pm$ 2.68} & 0.8499{\scriptsize $\pm$ 0.0464} & 0.2126{\scriptsize $\pm$ 0.0566} \\
\cmidrule(lr){1-6}
0 & 1.97 & 14.54 & 38.18{\scriptsize $\pm$ 1.82} & 0.9659{\scriptsize $\pm$ 0.0085} & 0.0232{\scriptsize $\pm$ 0.0112} \\
\textbf{1} & \textbf{31.89} & 17.54 & \textbf{39.19{\scriptsize $\pm$ 2.07}} & \textbf{0.9692{\scriptsize $\pm$ 0.0137}} & \textbf{0.0180{\scriptsize $\pm$ 0.0150}} \\
2 & 28.22 & 12.89 & 36.58{\scriptsize $\pm$ 2.26} & 0.9524{\scriptsize $\pm$ 0.0107} & 0.0320{\scriptsize $\pm$ 0.0080} \\
3 & 25.30 & 8.66 & 31.04{\scriptsize $\pm$ 2.51} & 0.8720{\scriptsize $\pm$ 0.0318} & 0.1729{\scriptsize $\pm$ 0.0423} \\
4 & 25.13 & 4.89 & 17.06{\scriptsize $\pm$ 1.84} & 0.3296{\scriptsize $\pm$ 0.0857} & 0.8279{\scriptsize $\pm$ 0.0754} \\
\bottomrule
\end{tabular}
\end{minipage}
\hfill
\begin{minipage}[t]{0.43\textwidth}
\centering STRAINER [\textbf{FFMLP}]\\[2pt]
\begin{tabular}{ccccc}
\toprule
sr(w) & sr(h) & PSNR (dB) $\uparrow$ & SSIM $\uparrow$ & LPIPS $\downarrow$ \\
\midrule
-- & -- & 28.66{\scriptsize $\pm$ 2.57} & 0.8189{\scriptsize $\pm$ 0.0521} & 0.2618{\scriptsize $\pm$ 0.0568} \\
\cmidrule(lr){1-5}
\textbf{66.77} & 33.32 & \textbf{30.35{\scriptsize $\pm$ 2.67}} & \textbf{0.8412{\scriptsize $\pm$ 0.0372}} & \textbf{0.2099{\scriptsize $\pm$ 0.0479}} \\
26.94 & 4.93 & 27.72{\scriptsize $\pm$ 2.48} & 0.7385{\scriptsize $\pm$ 0.0463} & 0.3565{\scriptsize $\pm$ 0.0557} \\
21.92 & 1.52 & 24.13{\scriptsize $\pm$ 2.00} & 0.5582{\scriptsize $\pm$ 0.0615} & 0.5640{\scriptsize $\pm$ 0.0552} \\
20.81 & 1.13 & 17.96{\scriptsize $\pm$ 2.19} & 0.3749{\scriptsize $\pm$ 0.0966} & 0.7886{\scriptsize $\pm$ 0.0634} \\
22.94 & 1.10 & 15.55{\scriptsize $\pm$ 2.12} & 0.4033{\scriptsize $\pm$ 0.1176} & 0.8747{\scriptsize $\pm$ 0.0858} \\
\bottomrule
\end{tabular}
\end{minipage}
\vspace{8pt}

\noindent\textit{OASIS $\to$ OASIS (in-dist)}\par
\vspace{2pt}
\begin{minipage}[t]{0.55\textwidth}
\centering STRAINER [\textbf{SIREN}]\\[2pt]
\begin{tabular}{cccccc}
\toprule
$\tau$ & sr(w) & sr(h) & PSNR (dB) $\uparrow$ & SSIM $\uparrow$ & LPIPS $\downarrow$ \\
\midrule
* & -- & -- & 48.98{\scriptsize $\pm$ 2.02} & 0.9939{\scriptsize $\pm$ 0.0066} & 0.0019{\scriptsize $\pm$ 0.0013} \\
\cmidrule(lr){1-6}
0 & 1.97 & 14.54 & 48.56{\scriptsize $\pm$ 1.93} & 0.9934{\scriptsize $\pm$ 0.0068} & 0.0027{\scriptsize $\pm$ 0.0008} \\
\textbf{1} & \textbf{31.89} & 17.54 & \textbf{49.17{\scriptsize $\pm$ 1.39}} & \textbf{0.9969{\scriptsize $\pm$ 0.0011}} & \textbf{0.0026{\scriptsize $\pm$ 0.0009}} \\
2 & 28.22 & 12.89 & 48.17{\scriptsize $\pm$ 1.49} & 0.9963{\scriptsize $\pm$ 0.0007} & 0.0035{\scriptsize $\pm$ 0.0009} \\
3 & 25.30 & 8.66 & 47.38{\scriptsize $\pm$ 2.15} & 0.9941{\scriptsize $\pm$ 0.0021} & 0.0032{\scriptsize $\pm$ 0.0011} \\
4 & 25.13 & 4.89 & 22.73{\scriptsize $\pm$ 1.78} & 0.5753{\scriptsize $\pm$ 0.0488} & 0.2032{\scriptsize $\pm$ 0.0192} \\
\bottomrule
\end{tabular}
\end{minipage}
\hfill
\begin{minipage}[t]{0.43\textwidth}
\centering STRAINER [\textbf{FFMLP}]\\[2pt]
\begin{tabular}{ccccc}
\toprule
sr(w) & sr(h) & PSNR (dB) $\uparrow$ & SSIM $\uparrow$ & LPIPS $\downarrow$ \\
\midrule
-- & -- & 44.11{\scriptsize $\pm$ 2.21} & 0.9908{\scriptsize $\pm$ 0.0019} & 0.0050{\scriptsize $\pm$ 0.0014} \\
\cmidrule(lr){1-5}
\textbf{66.77} & 33.32 & \textbf{44.04{\scriptsize $\pm$ 2.01}} & \textbf{0.9925{\scriptsize $\pm$ 0.0017}} & \textbf{0.0046{\scriptsize $\pm$ 0.0015}} \\
26.94 & 4.93 & 39.41{\scriptsize $\pm$ 2.19} & 0.9772{\scriptsize $\pm$ 0.0061} & 0.0164{\scriptsize $\pm$ 0.0053} \\
21.92 & 1.52 & 34.36{\scriptsize $\pm$ 2.10} & 0.9262{\scriptsize $\pm$ 0.0216} & 0.0579{\scriptsize $\pm$ 0.0141} \\
20.81 & 1.13 & 27.06{\scriptsize $\pm$ 2.35} & 0.7957{\scriptsize $\pm$ 0.0460} & 0.1444{\scriptsize $\pm$ 0.0245} \\
22.94 & 1.10 & 21.88{\scriptsize $\pm$ 1.81} & 0.6569{\scriptsize $\pm$ 0.0413} & 0.2039{\scriptsize $\pm$ 0.0168} \\
\bottomrule
\end{tabular}
\end{minipage}
\vspace{8pt}

\end{table*}

% \clearpage

% \newpage
% \input{checklist.tex}

\end{document}